\documentclass[journal]{vgtc}         

\ifpdf
  \pdfoutput=1\relax                   
  \usepackage{graphicx}                
  \DeclareGraphicsExtensions{.pdf,.png,.jpg,.jpeg} 
\else
  \usepackage{graphicx}                
  \DeclareGraphicsExtensions{.eps}     
\fi%

\graphicspath{{figures/}{pictures/}{images/}{./}} 

\usepackage{microtype}                 
\PassOptionsToPackage{warn}{textcomp}  
\usepackage{textcomp}                  
\usepackage{mathptmx}                  
\usepackage{times}                     
\usepackage{cite}                      
\usepackage{tabu}                      
\usepackage{booktabs} 
\usepackage{graphicx}
\usepackage{subcaption} 
\usepackage{multirow}
\usepackage{amsmath, amssymb}
\usepackage{hyperref}
\usepackage[capitalise]{cleveref}

\usepackage{color} 
\usepackage{colortbl}
\usepackage{cases}
\definecolor{light-gray}{gray}{0.6}
\definecolor{lavender}{rgb}{0.5,0.5,1.0}

\def\_{\rule{.3em}{.15ex}}      

\newcommand{\hot}[1]{{\color{black} #1}}
\newcommand{\comp}[1]{{\color{black} #1}}

\usepackage{lipsum}
\usepackage{fancyhdr}
\usepackage{tikz}

\definecolor{SomeColor}{RGB}{0,150,255}

\onlineid{2050}

\vgtccategory{Research}

\vgtcpapertype{Technique}

\title{Natural Language-Driven Viewpoint Navigation for Volume Exploration via Semantic Block Representation}

\author{Xuan Zhao, and Jun Tao, \textit{Member, IEEE}}
\authorfooter{
\item
 Xuan Zhao is with the School of Computer Science and Engineering, Sun Yat-sen University. E-mail: zhaox269@mail2.sysu.edu.cn.
\item
 Jun Tao is with the School of Computer Science and Engineering, Sun Yat-sen University, Guangdong Province Key Laboratory of Computational Science, and National Supercomputer Center in Guangzhou. E-mail: taoj23@mail.sysu.edu.cn. He is the corresponding author.
}

\abstract{
Exploring volumetric data is crucial for interpreting scientific datasets. However, selecting optimal viewpoints for effective navigation can be challenging, particularly for users without extensive domain expertise or familiarity with 3D navigation. In this paper, we propose a novel framework that leverages natural language interaction to enhance volumetric data exploration. Our approach encodes volumetric blocks to capture and differentiate underlying structures. It further incorporates a CLIP Score mechanism, which provides semantic information to the blocks to guide navigation. The navigation is empowered by a reinforcement learning framework that leverage these semantic cues to efficiently search for and identify desired viewpoints that align with the user’s intent. The selected viewpoints are evaluated using CLIP Score to ensure that they best reflect the user queries. By automating viewpoint selection, our method improves the efficiency of volumetric data navigation and enhances the interpretability of complex scientific phenomena.

} 

\keywords{Volume rendering,Viewpoint navigation, Natural language interaction}

\begin{document}


\firstsection{Introduction}

\maketitle
\comp{Volume visualization is crucial for understanding complex volumetric data. 
Direct volume rendering plays a key role in this process, where voxels are projected from 3D object space onto a 2D screen with assigned color and transparency. 
In addition, viewpoint selection will significantly impact the interpretation of structures~\cite{RUIZ2010351} as well, as the same geometry can appear drastically different from various viewing angles. 
A desired view can expose critical and unique structures, greatly enhancing the understanding of complex volumes.
In contrast, a poorly chosen viewpoint may fail to convey original 3D structures and introduce occlusions, limiting users' comprehension. 

Navigating and interpreting volumetric data is essential in scientific fields such as medical imaging~\cite{6820797}, geospatial analysis~\cite{8854316}, and flow visualization~\cite{tao2013VIs}. 
For instance, in medicine, doctors assess a patient's condition through CT scans, where abnormalities may only be visible from specific viewpoints.
Effective exploration of such data is crucial for extracting insights, yet traditional methods often rely on manual viewpoint selection. 
While this allows fine-grained inspection, it demands significant 3D navigation expertise and domain knowledge. 
As volumetric datasets grow in complexity and scale, there is a growing need for more intuitive, efficient, and automated exploration techniques.

Existing viewpoint selection techniques gradually transit from traditional ones based on geometric analysis and information theory to machine learning-based techniques.
Traditional techniques often emphasize visibility and occlusion reduction to reveal distinguished geometric structures, by assessing their complexities and information.
For example, unified information-theoretic frameworks have been proposed~\cite{tao2013VIs} to connect integral curves and viewpoint selection through information channels. 
Although mathematically rigorous, these methods rely on hand-crafted heuristics and often fail to account for subjective preferences.
In contrast, deep learning methods typically require large labeled datasets and learn from predefined criteria or expert selection results~\cite{shi2019cnns}.
While promising for multi-criteria recommendations, these methods still lack adaptability to diverse selection goals.
More importantly, both the traditional and existing learning-based techniques lack semantic understanding, limiting user-centered exploration. 
}

Recent advancements in artificial intelligence, particularly in vision-language models and reinforcement learning, have opened new avenues for intelligent volumetric data exploration. Inspired by these developments, we propose a novel framework that enables users to navigate volumetric datasets through natural language instructions, making 3D data interaction more intuitive and accessible. Our approach encodes volumetric data into structured semantic blocks, capturing meaningful features that guide viewpoint selection. \comp{A key innovation of our framework is the alignment of block representation, image representation, and textual descriptions, by integrating a CLIP-based~\cite{radford2021learning} semantic scoring mechanism. This empowers our system in understanding users' exploration goals in natural language.
Additionally, we employ reinforcement learning to optimize viewpoint selection dynamically, ensuring selected viewpoints accurately reflect the user's intent.}
The core contributions of our work are as follows:

\begin{itemize} 
\item  We introduce a natural language-driven approach for viewpoint specification, enabling intuitive interaction with volumetric data. This enhances the accessibility of volumetric exploration for users with diverse expertise levels.  
\item Our framework leverages CLIP to align textual descriptions with visual content and volumetric structures, ensuring that the selected viewpoints accurately capture the semantics of user queries, thereby enhancing navigation precision.
\item  We formulate viewpoint selection as a reinforcement learning problem, where the agent learns to optimize viewpoint choices based on semantic feedback, progressively refining its understanding of user intent.  
\item By segmenting the volume into semantically meaningful blocks, we introduce a structured representation that facilitates fine-grained feature extraction and viewpoint selection.  
\end{itemize}

\comp{In the following sections, we first review related work on viewpoint selection and AI-driven volumetric visualization. We then introduce our problem formulation and describe the proposed method in detail. This is followed by implementation details and experimental results that demonstrate the effectiveness of our approach. Finally, we analyze failure cases and discuss current limitations, with the hope that these insights will inform future research in AI-assisted exploration of scientific datasets.}

\section{Related Work}
\subsection{Viewpoint selection}

Viewpoint selection is a critical problem in 3D visualization, particularly in volume rendering~\cite{RUIZ2010351}, scientific visualization~\cite{tao2013VIs, Lee2011}, and medical imaging~\cite{preuhs2018viewpoint,ZHOU2023VI}. A well-chosen viewpoint can reveal important structures while minimizing occlusion and distortion, thereby improving both the clarity and interpretability of visual data.

Existing viewpoint selection methods can be broadly categorized into two main approaches: geometry-based optimization techniques and data-driven machine learning methods.
Early work on viewpoint selection primarily relied on geometric heuristics and optimization strategies. Bordoloi and Shen~\cite{Bordoloi2025VIS} proposed an information-theoretic approach for volume rendering, using voxel entropy to determine viewpoints that maximize information gain. Similarly, Takahashi et al.~\cite{takahashi2005feature} introduced a visibility-driven strategy aimed at optimizing camera positions to maintain the visibility of key features, especially in medical visualization contexts.
Another line of research focuses on perceptual importance and feature visibility. Viola et al.~\cite{Viola2005} developed an importance-driven viewpoint selection framework that assigns weights to visual features based on their relevance to the user’s task. By incorporating perceptual factors into the optimization process, their method enhances the interpretability and effectiveness of volumetric visualization.
Despite their contributions, geometric optimization methods often rely on hand-crafted heuristics and fixed objective functions. As a result, they may lack adaptability across varying visualization contexts, and they typically require manual parameter tuning. Furthermore, these approaches seldom account for user-specific preferences or high-level semantic intent.
With the rise of deep learning, data-driven methods have been explored to address these limitations. Shi and Tao~\cite{shi2019cnns} employed Convolutional Neural Networks (CNNs) to predict optimal viewpoints for volume visualization. Their model is trained on labeled datasets of preferred viewpoints, enabling it to infer suitable camera orientations for new volumetric data.

While machine learning-based methods improve generalization and adaptability, they also introduce new challenges. 
Training such models requires large-scale labeled datasets, which are often costly to produce in specialized domains. 
Moreover, many of these methods operate as black boxes, lacking interpretability and the ability to incorporate high-level user intent, such as semantic descriptions or task-specific goals, into the viewpoint selection process.

\subsection{Volumetric and 3D Data Representation}

\comp{In scientific visualization, volumetric data rendering has been widely studied. Early work established foundational techniques such as ray casting. Levoy~\cite{levoy1988display} introduced a core method for direct surface rendering from volume data, while Drebin et al.~\cite{brebin1998volume} proposed a volume rendering pipeline that visualizes scalar fields effectively through optical models and compositing. These methods primarily relied on grid-based volume discretizations.
With increasing data resolution and size, multiscale representations became essential for efficient rendering. LaMar et al.~\cite{lamar1999multiresolution} introduced hierarchical multiresolution structures for interactive visualization of large volumes. Beyer et al.~\cite{beyer2015state} surveyed GPU-based methods, highlighting scalable data structures and algorithms for large-scale volume rendering.
Driven by 3D recognition tasks, structured volumetric representations like voxel grids and octrees gained traction in data-driven methods. VoxNet~\cite{maturana2015voxnet} integrated voxel grids with 3D CNNs for real-time object recognition. Addressing voxel inefficiencies, Wang et al.~\cite{wang2017cnn} developed O-CNN, which combines octrees with CNNs for more efficient deep inference.
Recently, implicit neural representations have emerged as flexible, memory-efficient alternatives. DeepSDF~\cite{park2019deepsdf} models 3D shapes as continuous signed distance fields using neural networks. Occupancy Networks~\cite{mescheder2019occupancy} predict point-wise occupancy in 3D space, enabling compact, resolution-free representations. Notably, NeRF~\cite{mildenhall2021nerf} encodes volumetric radiance fields with neural networks, enabling high-quality view synthesis and inspiring extensive follow-up research.}


\subsection{Vision-Language Representation}
The integration of vision and language tasks has been a longstanding area of research, focusing on aligning textual descriptions with visual content. Initial efforts primarily used hand-crafted features to bridge the gap between visual and textual modalities. Barnard et al.\cite{barnard2003matching} and Duygulu et al.\cite{duygulu2002object} introduced pioneering probabilistic models to match regions in images with corresponding words, utilizing co-occurrence statistics to establish associations between the two modalities. These approaches laid crucial groundwork in understanding cross-modal correspondences.

The advent of deep learning brought about significant advancements in vision-language tasks, leveraging the power of neural networks to learn joint representations. Notably, Karpathy and Fei-Fei~\cite{karpathy2015deep} proposed a deep neural network model that seamlessly connects image regions with sentence fragments through a multimodal embedding space, facilitating tasks like image description and captioning. Concurrently, Mao et al.~\cite{mao2014deep} developed the NIC (Neural Image Captioning) model, integrating a convolutional neural network with a recurrent neural network to encode images and generate natural language descriptions, marking a pivotal development in generating coherent textual descriptions for images.


\comp{More recently, Vision-Language Pre-training (VLP) has emerged as a powerful paradigm, using large-scale data to learn generalized cross-modal representations. Models like LXMERT~\cite{tan2019lxmert} and VilBERT~\cite{lu2019vilbert}, inspired by BERT~\cite{devlin2018bert}, extended masked language modeling to visual domains. CLIP~\cite{radford2021learning} further advanced the field with contrastive learning to align vision and text embeddings, excelling in zero-shot settings.
Tasks such as vision-language navigation and grounding highlight the need to comprehend spatial and contextual language. Anderson et al.~\cite{anderson2018vision} introduced the VLN task, where agents interpret navigation instructions in 3D environments. Tan et al.~\cite{tan2019learning} extended this with specialized pre-training techniques for navigation, improving task generalization.}

Transformers have become a backbone architecture for multimodal tasks, with adaptations like the Vision Transformer (ViT)\cite{dosovitskiy2020image} influencing the design of unified transformers for vision-language tasks. UNITER\cite{chen2020uniter} showcased a unified transformer architecture that learns vision-language joint embeddings using extensive pre-training strategies, leading to state-of-the-art results in tasks like Visual Question Answering (VQA) and image-text matching. Further, VLMo~\cite{bao2022vlmo} unified text, vision, and vision-language pre-training under a shared model, highlighting ongoing efforts to achieve comprehensive multimodal understanding through unified frameworks.

\begin{figure*}[ht]
    \centering
    \includegraphics[width=1\textwidth]{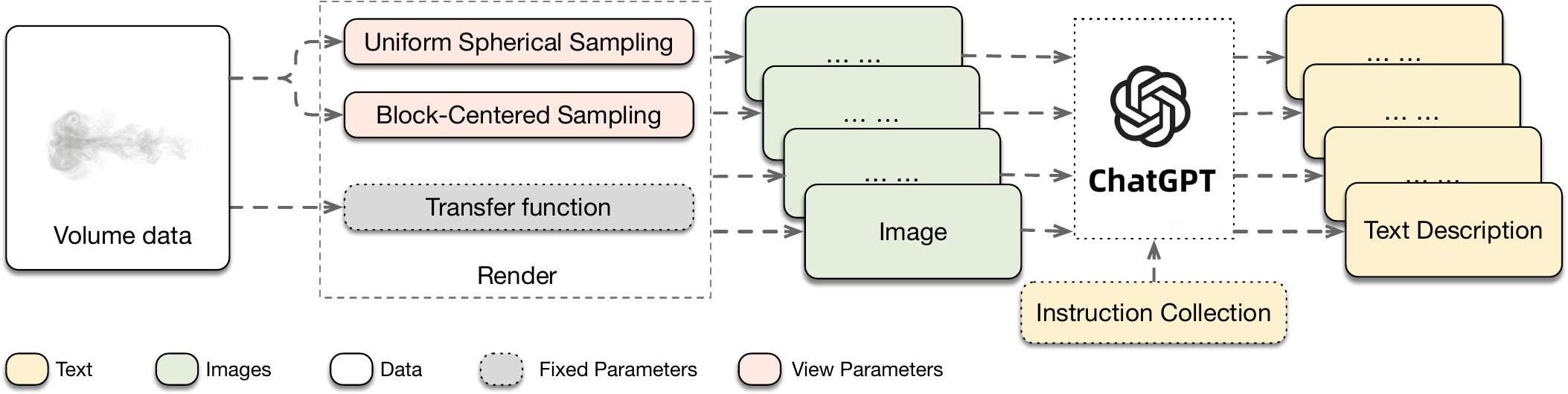} 
    \caption{Overview of the automatic text annotation process using ChatGPT. The volumetric data is first rendered with an appropriate transfer function. Two sampling strategies, Uniform Spherical Sampling and Block-Centered Sampling, are applied to extract relevant viewpoints. The sampled images are then processed by ChatGPT to generate textual descriptions, guided by an instruction collection mechanism that refines the annotation process.}
    \label{GPTtext}
\end{figure*}

\subsection{Reinforcement Learning for Visual Tasks}

Reinforcement Learning (RL) has been widely applied to visual tasks, enabling agents to make sequential decisions by learning from environmental interactions. Foundational work by Sutton and Barto~\cite{sutton1998reinforcement} laid the groundwork for RL algorithms, which have since evolved to incorporate high-dimensional visual data.

The integration of deep learning with RL, commonly known as Deep Reinforcement Learning (DRL), has driven significant progress. A landmark example is the Deep Q-Network (DQN) by Mnih et al. \cite{mnih2015human}, which combined CNNs with Q-learning to play Atari games directly from raw pixel inputs.
This work demonstrated the feasibility of end-to-end visual decision-making, inspiring numerous DRL extensions.
Proximal Policy Optimization (PPO), introduced by Schulman et al.~\cite{schulman2017proximal}, has emerged as a robust and sample-efficient policy gradient method. Its clipped surrogate objective facilitates stable learning and makes PPO particularly effective for continuous control and high-dimensional visual tasks. Owing to its balance of simplicity and performance, PPO has become a popular baseline in visual RL benchmarks.
RL has also shown success in visual navigation and control. Zhu et al.~\cite{zhu2017target} applied RL to target-driven navigation, while Lillicrap et al.~\cite{lillicrap2015continuous} introduced Deep Deterministic Policy Gradient (DDPG) for continuous action control using visual inputs. These methods enable agents to interact with their environments using visual cues, pushing forward the capabilities of embodied AI.
Addressing exploration and sample efficiency, methods like Rainbow~\cite{hessel2018rainbow} combine multiple improvements over DQN to accelerate learning, while curiosity-driven approaches~\cite{pathak2017curiosity} promote intrinsic motivation for agents to explore visual scenes more effectively.
Recent research emphasizes RL in rich and dynamic environments. 
Benchmarks such as AI Habitat~\cite{savva2019habitat} and ViZDoom~\cite{kempka2016vizdoom} provide realistic testbeds for vision-based agents. 

Together, these advancements demonstrate the potential of reinforcement learning as a foundation for semantic, visual interaction in environments with high-dimensional input spaces, a capability essential for tasks such as language-guided volumetric exploration.

\section{Our Approach}

\subsection{Overview}
We propose a natural language-driven framework for automating viewpoint selection, leveraging volumetric block encoding, CLIP-based semantic guidance, and reinforcement learning. The framework consists of four main components: data preparation, CLIP fine-tuning, semantic block encoding, and reinforcement learning based on the encoded features.
\emph{Data preparation} involves constructing paired image-text datasets for vision-language training. Volume data is rendered from uniformly sampled viewpoints by subdividing an icosahedron centered on the volume, and ChatGPT is employed to generate corresponding textual descriptions.
\emph{CLIP fine-tuning} enhances the model's sensitivity to 3D structural features, which are often underrepresented in CLIP models pretrained on natural images. This is achieved using the constructed image-text dataset.
\emph{Semantic block encoding} captures the localized semantic content of volumetric features. The volume is evenly partitioned into blocks, which are first encoded using an autoencoder and then aligned with text embeddings via the paired dataset. During reinforcement learning, this aligned semantic representation replaces CLIP's original image embeddings to improve training efficiency.
Finally, viewpoint selection is formulated as a \emph{reinforcement learning} task, where the agent learns to identify informative viewpoints guided by a CLIP-based semantic reward. By integrating the semantic encoding with CLIP feedback, our framework facilitates efficient exploration of complex volumetric structures and enhances interpretability in scientific visualization.
The following sections detail each component of the proposed approach.


\subsection{Data Preparation}
\subsubsection{Volume Rendering and Viewpoint Sampling}
\label{sec:view-sample}
This section describes the sampling strategies used to construct the image dataset, which forms the foundation for fine-tuning the CLIP model. 
The viewpoints are selected to ensure comprehensive coverage of the volumetric space, facilitating both global context understanding and detailed local structure analysis.

\textbf{Uniform Spherical Sampling.}
\hot{The viewpoints $V = \{v_i\}_{i=1}^{N}$ are evenly sampled on a sphere enclosing the volume, by recursively subdividing a regular icosahedron. At each subdivision level \( k \), every face of the triangular mesh is divided into four smaller triangles by bisecting its three edges and connecting the midpoints. The centers of triangles are used as sample viewpoints, leading to $N = 20 \times 4^k$ viewpoints at level $k$. In our experiments, we set \( k = 2 \), resulting into \( N = 320 \) viewpoints.}




All cameras are oriented such that their optical axes point toward the volume center, ensuring uniform coverage and reducing viewpoint bias. This configuration allows for systematic observation from multiple directions, promoting global feature awareness during training.

\textbf{Block-Centered Sampling.}
To improve viewpoint diversity and enhance sensitivity to localized structures, we introduce a block-centered sampling strategy. Instead of directing all viewpoints toward the global volume center, this approach reorients cameras toward the geometric centers of individual volumetric blocks.
\hot{Formally, given a set of blocks \( B = \{b_j\}_{j=1}^{M} \), each block \( b_j \) has a center denoted as $\mathbf{c}_j$.
For a given viewpoint \( v_i \), the camera is reoriented to face a specific block center \( c_j \), resulting in a direction vector:

\begin{equation}
\mathbf{d}_{ij} = \frac{\mathbf{c}_j - \mathbf{v}_i}{\| \mathbf{c}_j - \mathbf{v}_i \|}.
\end{equation}
}

\noindent This formulation ensures that local structures within blocks are more prominently captured, improving the model’s ability to encode fine-grained spatial features.

By combining uniform spherical sampling and block-centered sampling, our dataset construction strategy achieves a balance between global consistency and local adaptability. This dual-viewpoint approach supports comprehensive volumetric understanding and improves the semantic quality of CLIP-based alignment.

\subsubsection{Automatic Text Annotation via ChatGPT}
Based on the rendered image dataset, we generate a corresponding set of textual annotations to describe the visual content of each viewpoint. 
These descriptions are essential for establishing semantic alignment between visual and linguistic representations, enabling the CLIP model to learn meaningful cross-modal associations during training.
To ensure the quality and relevance of the generated captions, we leverage a GPT-based language model (ChatGPT) known for its advanced natural language generation capabilities. This allows us to produce textual descriptions that are both detailed and semantically faithful to the visual features of each rendered image.

As illustrated in Fig.~\ref{GPTtext}, the annotation process begins with a set of rendered volumetric images, denoted as \( I = \{ I_i \}_{i=1}^{N} \). For each image \( I_i \), we use the ChatGPT API to generate a descriptive caption. 
A carefully designed prompt is provided alongside each image, instructing the model to focus on key structural elements visible in that specific viewpoint. 
The resulting textual annotation \( T_i \) is formulated as:

\begin{equation}
    T_i = \text{ChatGPT}(I_i, \text{prompt}),
\end{equation}
where \text{ChatGPT}($\cdot$) denotes the API function, and \text{prompt} specifies the desired format and level of detail in the output description.
To ensure high-quality annotations, we carefully design the prompt to encourage descriptions that capture key structural features, variations in density, and prominent patterns within the volumetric data. An example of such a prompt is:
\begin{quote}
    ``Describe the fish's appearance, including its skeletal structure and notable features, based on the given perspective."
\end{quote}
By following this procedure, we obtain a dataset of paired image-text samples $D = \{(I_i, T_i)\}_{i=1}^{N}$, which is used for CLIP fine-tuning.


\begin{figure}[h]
    \centering
    \includegraphics[width=0.45\textwidth]{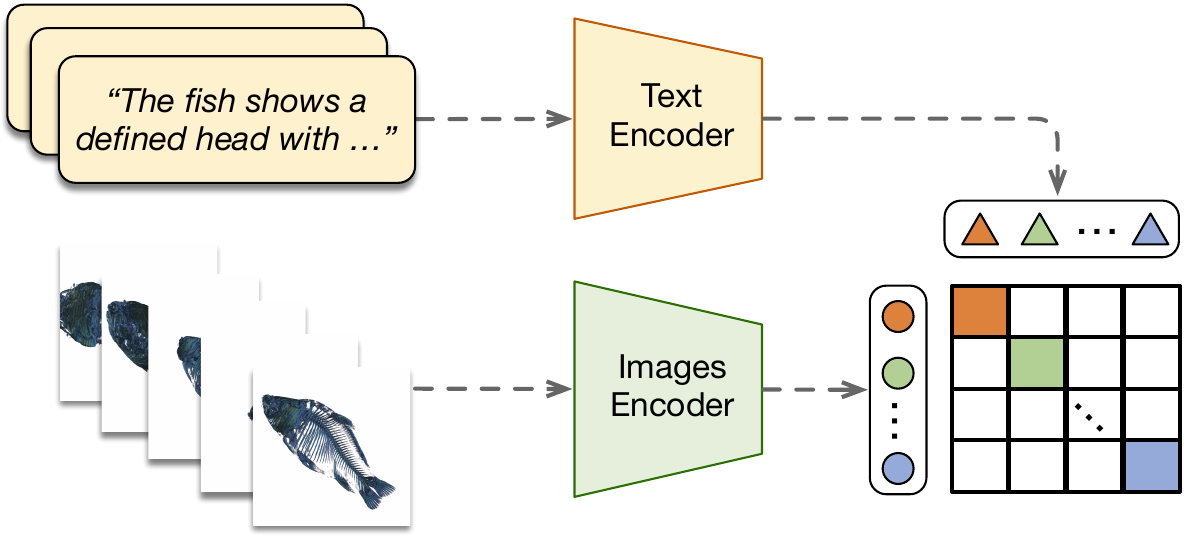} 
    \caption{CLIP-based fine-tuning architecture tailored for volumetric data exploration. Rendered images and their GPT-generated textual descriptions are encoded into a shared embedding space using CLIP’s vision and text encoders. This embedding alignment is used as the semantic backbone for viewpoint selection based on natural language instructions.}
    \label{finetuning clip}
\end{figure}

\subsection{CLIP Fine-tuning for Volumetric Data}
\hot{We adapt the pretrained CLIP model to better understand volumetric data by fine-tuning it on a specialized dataset of rendered volumetric images paired with descriptive texts. This fine-tuning process leverages contrastive learning to improve alignment between image and text embeddings, enabling the model to more effectively capture features specific to volumetric representations (see Fig.~\ref{finetuning clip}).}

Our dataset consists of paired image-text samples $D = \{(I_i, T_i)\}_{i=1}^{N}$, where $I_i$ is a rendered image capturing volumetric data and $T_i$ is its corresponding description generated by ChatGPT.
We utilize the original CLIP loss function, which is a symmetric cross-entropy loss over cosine similarities. Given a batch of $P$ image-text pairs, the loss for \emph{image-to-text} matching is computed as:
\begin{equation}
L_{i2t} = - \frac{1}{P} \sum_{i=1}^{P} \log \frac{\exp(\text{sim}(z_i, t_i) / \tau)}{\sum_{j=1}^{P} \exp(\text{sim}(z_i, t_j) / \tau)}, 
\end{equation}
where $\text{sim}(z_i, t_i) = \langle z_i, t_i \rangle / (\|z_i\| \|t_i\|)$ represents the cosine similarity between the normalized image embedding $z_i = \text{CLIP}_\text{img}(I_i)$ and text embedding $t_i = \text{CLIP}_\text{text}(T_i)$. \hot{The temperature parameter $\tau$ is learned during training. Similarly, the loss for \emph{text-to-image} matching is:}

\begin{equation}
L_{t2i} = - \frac{1}{P} \sum_{i=1}^{P} \log \frac{\exp(\text{sim}(t_i, z_i) / \tau)}{\sum_{j=1}^{P} \exp(\text{sim}(t_i, z_j) / \tau)}.
\end{equation}

The overall contrastive loss is the sum of these two components:

\begin{equation}
L_{\text{contrastive}} = \frac{1}{2} (L_{i2t} + L_{t2i}).
\end{equation}

\begin{figure}[h]
    \centering
    \includegraphics[width=0.45\textwidth]{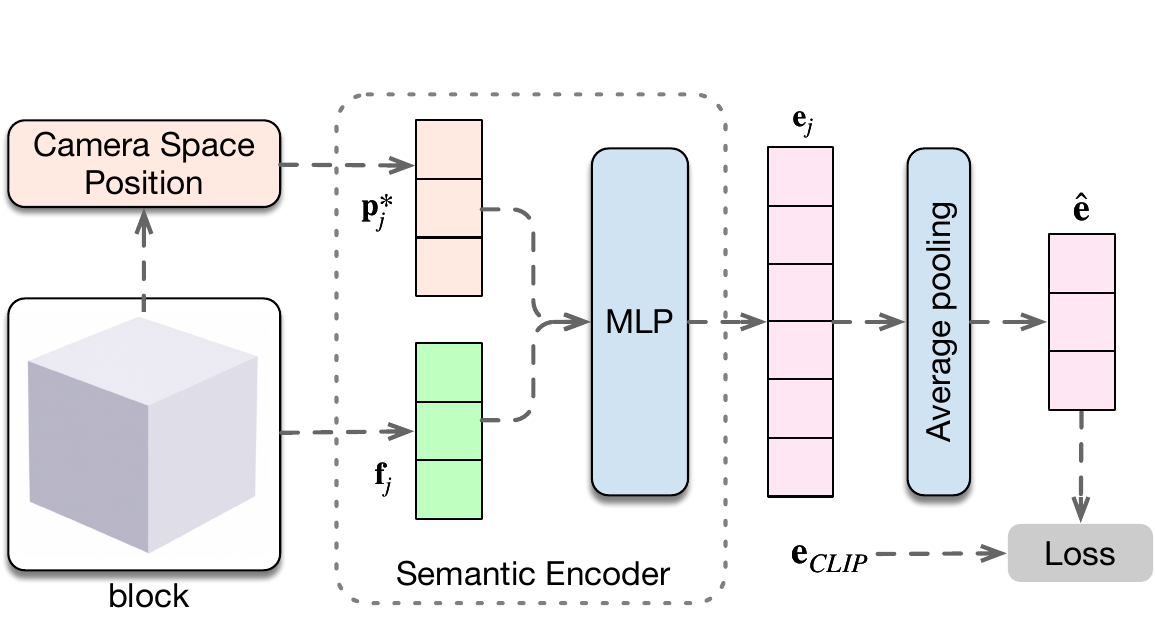} 
    \caption{\hot{An illustration of the semantic block encoding that converts volumetric feature encoding $f_j$ and camera space position $p_j^*$ into a semantic representation $e_j$.}}
    \label{bolck mlp}
\end{figure}

\subsection{Semantic Block Encoding}
While fine-tuned CLIP captures global semantic features in volumetric images, it may be less sensitive to local structural details. This limits the exploration of localized features, such as turbulence patterns in fluid dynamics or anomalies in medical scans. 

To enhance the sensitivity of viewpoint selection to local features, we introduce a block-based encoding mechanism that models the relationship between viewpoints and volumetric structures. The volumetric data is partitioned into a set of blocks, as previously defined in the viewpoint sampling process. Each block serves as a basic unit for carrying semantic information of local structures.

 


To fully capture the spatial relationship between viewpoints and local structures, we introduce a camera-space encoding mechanism. This encoding represents the position of each visible block relative to the camera, enabling the model to differentiate between blocks based on viewpoint-dependent visibility.
\hot{For a given viewpoint \( v_i \), a block $b_j$ is visible if the block intersects with the corresponding view frustum. All visible blocks under $v_i$ is further denoted as a set:}

\begin{equation}
B_i = \{ b_j \mid b_j \text{ is visible from } v_i \}.
\end{equation}

\hot{The semantic encoding of a visible block $b_j$ is illustrated in Fig.~\ref{bolck mlp}.}
The block \( b_j \) is first projected into the camera coordinate system, obtaining a position vector:

\begin{equation}
\mathbf{p}_{ij} = \text{Proj}(b_j, v_i),
\end{equation}

\noindent where \( \text{Proj}(\cdot) \) maps the global coordinates of block $b_j$ into the local camera space of viewpoint $v_i$.

By encoding each block’s position relative to the camera, we obtain a structured representation of local structures in different views. Each block \( b_j \) receives a position vector \( \mathbf{p}_{ij} \) in the camera space, forming a viewpoint-aware encoding of local structures. This information will later be used in the embedding learning process to improve feature sensitivity to local structures.



The camera-space encoding provides spatial priors for training, ensuring that viewpoint selection is guided not only by global semantics but also by localized volumetric structures.
To integrate volumetric block features and spatial encoding, we adopt a specialized convolution-fully connected hybrid architecture volume autoencoder. 
This model encodes local volumetric structures and their spatial positions, producing embeddings that align with CLIP's visual representations.

For each volumetric block \( b_j \), we employ a 3D convolutional autoencoder to extract compact and meaningful local visual features. Given an input volumetric block \( b_j \), the encoded feature representation is obtained as:

\begin{equation}
\mathbf{f}_j = \text{CNN}_{3D}(b_j),
\end{equation}

\noindent where \( \mathbf{f}_j \in \mathbb{R}^{d_f} \) is the extracted feature vector of the block, with \( d_f \) denoting the dimensionality of the feature representation.

To incorporate spatial context, the relative camera-space position \( \mathbf{p}_{ij} \) of each block \( b_j \) under viewpoint \( v_i \) is encoded using a multi-layer perceptron:

\begin{equation}
\mathbf{p}^*_j = \text{MLP}_{\text{pos}}(\mathbf{p}_{ij}),
\end{equation}

\noindent where \( \mathbf{p}^*_j \in \mathbb{R}^{d_p} \) represents the transformed positional feature, with \( d_p \) denoting the dimensionality of the positional feature representation.

\hot{
For each visible volumetric block \( b_j \), its visual feature \( \mathbf{f}_j \) and positional feature \( \mathbf{p}^*_j \) are concatenated and fed into a fusion network to obtain a semantic embedding:

\begin{equation}
\mathbf{e}_j = \text{MLP}_{\text{fusion}}([\mathbf{f}_j ; \mathbf{p}^*_j]),
\end{equation}
\noindent where \( \mathbf{e}_j \in \mathbb{R}^{d_e} \) is a per-block representation capturing both appearance and spatial context. To obtain a global descriptor, the per-block representations of all visible blocks are combined using average pooling, resulting into a single vector \( \hat{\mathbf{e}} \in \mathbb{R}^{768} \). 
}






\textbf{Training Objective.}  
\hot{
To ensure that the learned volumetric embeddings approximate semantically meaningful CLIP features, we train the block encoder using a cosine embedding loss defined as:

\begin{equation}
\mathcal{L}_{\text{cosine}} = 1 - \cos(\hat{\mathbf{e}}, \mathbf{e}_{\text{CLIP}}),
\end{equation} 
where $\hat{\mathbf{e}}$ is the learned embedding and \(\mathbf{e}_{\text{CLIP}}\) is the CLIP embedding derived from the rendered image. Minimizing this loss increases the cosine similarity between $\hat{\mathbf{e}}$ and $\mathbf{e}_{\text{CLIP}}$ and aligns the learned embedding space with CLIP's semantic feature space.
}

\begin{figure}[h]
    \centering
    \includegraphics[width=0.45\textwidth]{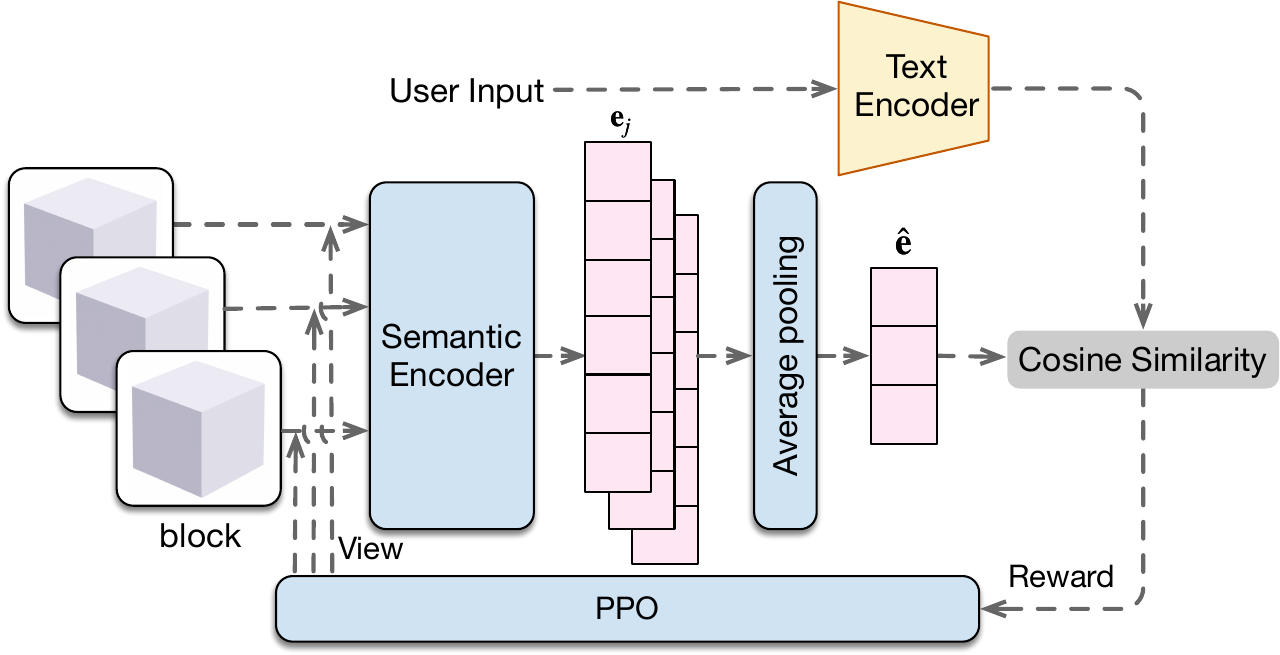}
    \caption{ \hot{Architecture of the PPO-based reinforcement learning framework for natural language-driven viewpoint selection. User instructions are processed through a text encoder to obtain semantic embeddings, which are compared with the predicted visual embeddings using cosine similarity. The resulting reward is used to train the PPO agent to iteratively refine camera viewpoints that best match the user’s intent.}}
    \label{PPO}
\end{figure}

\subsection{Reinforcement Learning for Optimal Viewpoint Selection}

To automatically identify viewpoints that best align with a given textual description, we formulate the viewpoint selection task as an RL problem. Specifically, our proposed RL framework integrates the semantic encoder model with a fine-tuned CLIP model, iteratively optimizing camera parameters to maximize semantic alignment. The virtual camera viewpoint is parameterized by a quaternion for orientation and a scalar depth parameter for distance control, as shown in Fig.~\ref{PPO}.


\textbf{Environment Formulation}
We define our environment following the OpenAI Gym interface paradigm. 
\hot{This state is input to a semantic encoder that processes volumetric blocks visible from the current viewpoint to produce per-block embeddings. These are aggregated (e.g., by average pooling) into a global semantic embedding \(\hat{\mathbf{e}}\) representing the rendered view.}

\textbf{State and Action Space}
Both the state and action spaces are defined as continuous vectors in $\mathcal{R}^5$, comprising a four-dimensional quaternion for orientation and an additional scalar $d$ representing the camera depth (distance from the object center). Formally:
\begin{equation}
  \mathcal{S}, \mathcal{A} \subseteq [-1, 1]^4 \times [d_{\text{min}}, d_{\text{max}}].
\end{equation}
At each step, the RL agent proposes incremental adjustments to the current quaternion and depth value. The quaternion is subsequently normalized, and the depth value is clamped within predefined limits to ensure valid camera positioning.


\textbf{Reward Function}
\hot{User natural language instructions are embedded by a fine-tuned CLIP text encoder to produce a semantic vector \(\mathbf{g}\). The reward at each step is computed as the cosine similarity between the volumetric feature \(\hat{\mathbf{e}}\) of the current viewpoint and the text embedding \(\mathbf{g}\):

\begin{equation}
r_t = \cos(\hat{\mathbf{e}}, \mathbf{g}) = \frac{\hat{\mathbf{e}} \cdot \mathbf{g}}{\|\hat{\mathbf{e}}\| \|\mathbf{g}\|},
\end{equation}
guiding the agent to select viewpoints that best align with the prompt’s semantic content.}


\textbf{PPO Training}
We adopt Proximal Policy Optimization (PPO) to train the viewpoint selection agent. PPO is well-suited for this task as it stabilizes policy updates while maintaining sample efficiency. The policy network \( \pi_{\theta}(a_t \mid s_t) \) is optimized using the clipped surrogate objective:

\hot{
\begin{equation}
L^{\text{PPO}}(\theta) = \mathbb{E}_t \left[ \min\left( \rho_t(\theta) A_t, \text{clip}(\rho_t(\theta), 1-\epsilon, 1+\epsilon) A_t \right) \right],
\end{equation}
where \( \rho_t(\theta)  \)is the policy probability ratio between the new and old policies,} \( A_t \) is the advantage estimate, and \( \epsilon \) is a clipping threshold to prevent excessively large updates.

\comp{During training, our reinforcement learning framework iteratively refines the viewpoint-selection policy via interactions with the environment. At each timestep \( t \), the current volumetric scene and camera parameters are encoded into a state vector \( s_t \). The policy outputs an action \( a_t \) that incrementally adjusts the camera orientation and depth, updating the viewpoint to \( s_{t+1} \).
The environment then renders a new view, and the reward \( r_t \) is computed based on the semantic similarity between the rendered embedding and the textual prompt embedding, measured by the fine-tuned CLIP model. The collected transitions, actions, and rewards update the PPO policy network.
Through repeated optimization, the agent learns to select viewpoints that maximize semantic alignment with user instructions, enabling more intuitive and effective exploration of volumetric data guided by natural language.}

\section{Evaluation}

\subsection{Quantitative Results}
To comprehensively evaluate the effectiveness of the proposed framework, we conduct experiments across multiple volumetric datasets and analyze key hyperparameters. This section begins by introducing the datasets used for training and evaluation, followed by detailed implementation specifics, including CLIP fine-tuning and viewpoint selection. We then include a training time analysis to demonstrate computational efficiency, assess the impact of different reward functions on viewpoint optimization, and finally evaluate the improvements brought by CLIP fine-tuning through quantitative performance comparisons.

\textbf{Datasets.}  
We evaluated our framework using three distinct volumetric datasets from diverse domains. The first, the Carp Fish dataset, consists of CT-scanned volumetric data of a fish and is used primarily to test anatomical viewpoint navigation. The second, the Skull dataset, is derived from a rotational C-arm X-ray scan of a human skull phantom, representing a typical medical imaging scenario. The third, the Argon Bubble dataset, represents a fluid dynamics simulation and is used to assess the system’s performance on physically-based structures. These datasets collectively offer a comprehensive basis for evaluating the generalizability and robustness of the proposed method.

\begin{table}[h]
\centering
\caption{Dataset specifications and block configurations.}
\small 
\begin{tabular}{l c c c}
\hline
\textbf{Dataset} & \textbf{Dimensions} & \textbf{Number of Blocks} & \textbf{Block Size} \\
\hline
Carp Fish & 256 $\times$ 256 $\times$ 512 & 64 & 64 $\times$ 64 $\times$ 128 \\
Skull & 256 $\times$ 256 $\times$ 256 & 64 & 64 $\times$ 64 $\times$ 64 \\
Argon Bubble & 256 $\times$ 256 $\times$ 640 & 64 & 64 $\times$ 64 $\times$ 160 \\
\hline
\end{tabular}
\label{tab:dataset_specifications}
\end{table}

\textbf{Implementation Details.}
To effectively adapt CLIP for volumetric data, we carefully selected hyperparameters based on empirical evaluation. Key parameters include the number of sampled viewpoints, volumetric block configurations, and model training settings such as batch size, learning rate, and the use of contrastive loss.

Experiments were conducted on all three datasets, each with varying spatial resolutions. Accordingly, we adjusted dataset-specific parameters to suit the characteristics of each volume, rather than using a one-size-fits-all configuration. Table~\ref{tab:dataset_specifications} summarizes the dataset specifications and block configurations. In contrast, Table~\ref{tab:hyperparameters} lists global hyperparameters used consistently across all training sessions.

\hot{
For all datasets, we used both Uniform Spherical Sampling and Block-Centered Sampling (Sec. \ref{sec:view-sample}) to generate a diverse set of viewpoints. Specifically, using recursive icosahedral subdivision at level $k = 2$, we obtained 320 uniformly distributed viewpoints over the spherical surface. 
To further emphasize local structures, we applied block-centered sampling by randomly selecting 10 volumetric blocks and pairing each with 10 randomly chosen directions from the uniform set, resulting in an additional 100 targeted viewpoints. In total, we utilized 420 distinct viewpoints per dataset in our experiments.
}

\begin{table}[h]
\centering
\caption{Global hyperparameters used for CLIP fine-tuning.}
\begin{tabular}{l c}
\hline
\textbf{Parameter} & \textbf{Value} \\
\hline
Viewpoint Sampling Level ($k$) & 2 \\
Number of Viewpoints ($N$) & 420 \\
Batch Size & 128 \\
Learning Rate & $5 \times 10^{-5}$ \\
Optimizer & AdamW \\
Training Epochs & 100 \\
\hline
\end{tabular}
\label{tab:hyperparameters}
\end{table}

CLIP fine-tuning was performed using a batch size of 128 and a learning rate of $5 \times 10^{-5}$, optimized via the AdamW optimizer. Training was conducted for 100 epochs using contrastive learning objectives. These settings ensured effective alignment between visual embeddings and textual descriptions across the diverse volumetric inputs.

\begin{table}[h]
\centering
\caption{\hot{Training time for each component of our framework.}}
\begin{tabular}{l c}
\hline
\textbf{Component} & \textbf{Time} \\
\hline
Viewpoint Sampling & 23.02 sec\\
CLIP fine-tuning & 92.84 min \\
Volumetric block semantic encoding & 48.15 min\\
PPO-based viewpoint optimization & 21.38 min \\
\hline
\end{tabular}
\label{tab:training_time}
\end{table}

\hot{
\textbf{Training Time Analysis.} 
To understand the computational efficiency of our framework, we report the average runtime of each major component across the three datasets, as summarized in Table~\ref{tab:training_time}.
All results are measured on a single NVIDIA RTX 4090 GPU. The initial viewpoint sampling comprises two complementary strategies to generate 420 samples for each dataset. This sampling procedure completes in 23.02 seconds.
Fine-tuning the CLIP model on our paired image-text dataset, which consists of one textual description per rendered image, takes approximately 92.84 minutes. In total, we use 420 image-text pairs for training. Training the volumetric block encoder to approximate CLIP embeddings from local volume representations takes approximately 48.15 minutes. Finally, the PPO-based viewpoint policy optimization module converges in about 21.38 minutes. These results demonstrate that our model can be trained efficiently in hours for each dataset.
}

\begin{figure}[h]
    \centering
    \includegraphics[width=0.45\textwidth]{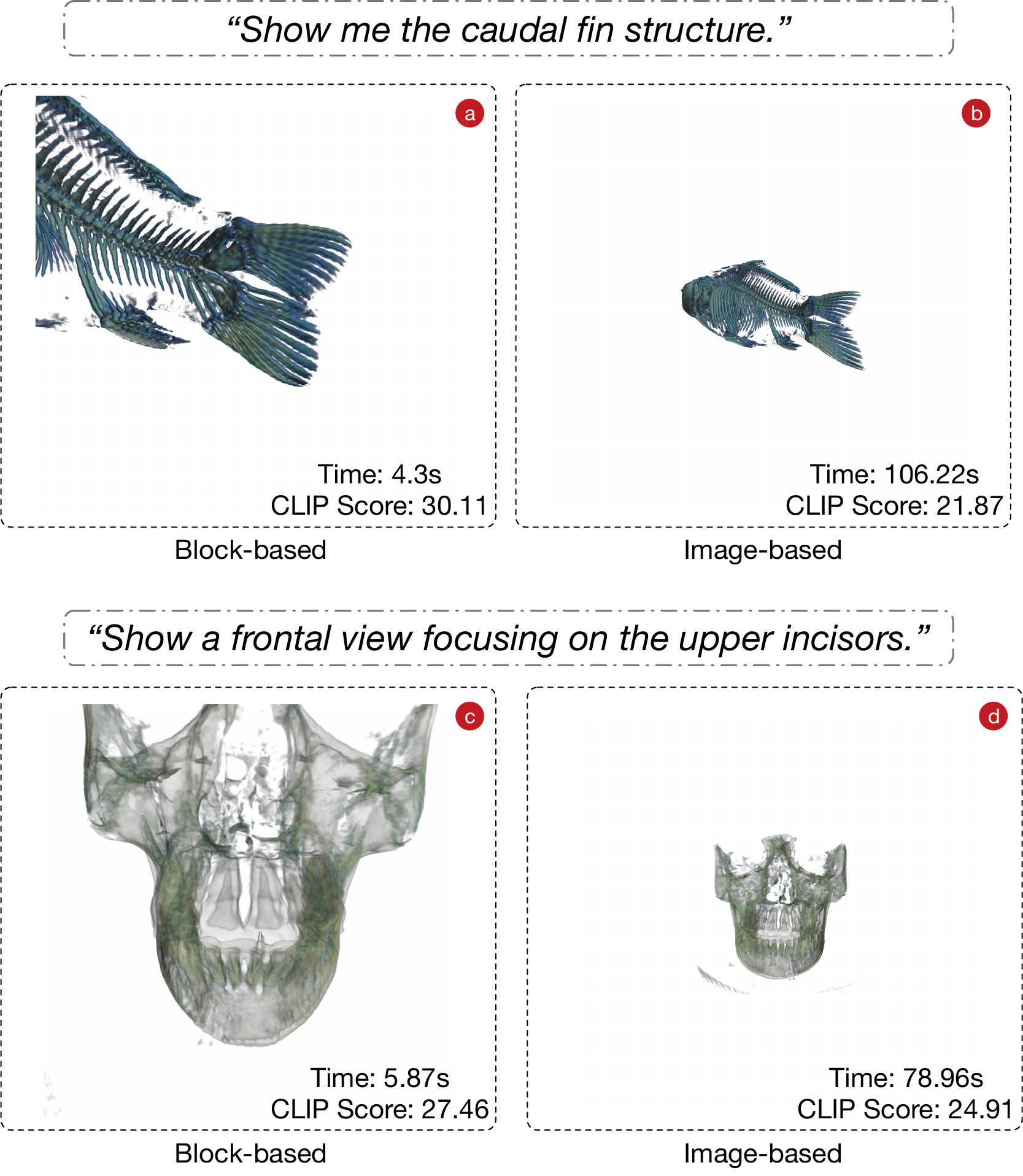}
    \caption{\hot{Comparison of viewpoint selection using block-based reward and image-based reward across different prompts and datasets. 
The first row shows results for the Carp Fish dataset under the prompt: ``\textit{Show me the caudal fin structure.}'' The second row shows results for the Skull dataset under the prompt: ``\textit{Show a frontal view focusing on the upper incisors.}'' 
In both cases, block-based reward enables more precise localization of semantic targets, while full-image reward often yields less focused or misaligned viewpoints.}
}
    \label{fig:reward_ablation_cases}
\end{figure}

\hot{
\textbf{Impact of Reward Evaluation Method.}
To investigate the influence of reward design on viewpoint selection, we compare our proposed block-based reward against a baseline that uses full image embeddings computed from rendered views.
Fig.~\ref{fig:reward_ablation_cases} shows qualitative results and performance metrics under two prompts from distinct datasets: the Carp Fish dataset with the prompt \textit{``Show me the caudal fin structure.''} (top row), and the Skull dataset with the prompt \textit{``Show a frontal view focusing on the upper incisors.''} (bottom row). For each dataset, the left column illustrates results obtained using the block-based reward, while the right column shows the corresponding outcomes with the image-based reward.

The block-based reward consistently guides the system to more precise and semantically aligned viewpoints, clearly revealing the targeted anatomical structures. In contrast, the image-based reward often yields less focused or misaligned views that lack sufficient spatial specificity for fine-grained localization.
Each subfigure includes inference time and CLIP Score annotations at the bottom right, indicating that the block-based reward achieves significantly faster convergence, around 4 to 6 seconds, while the image-based reward requires over 78 to 106 seconds. This efficiency gain arises from the localized gradient feedback in block-level representations, which better captures the alignment between visual features and user intent.
}

\begin{table}[h]
\centering
\caption{Performance of the CLIP model on three datasets before and after fine-tuning.}
\small 
\begin{tabular}{l c c}
\hline
\textbf{Dataset} & \textbf{Before Fine-tuning} & \textbf{After Fine-tuning} \\
\hline
Carp Fish & 18.16 & \textbf{28.32} \\
Skull & 20.70 & \textbf{26.01} \\
Argon Bubble  & 13.49 & \textbf{23.42} \\
\hline
\end{tabular}
\label{tab:clip_finetuning_scores}
\end{table}

\begin{figure*}[th]
    \centering
    \includegraphics[width=1.0\textwidth]{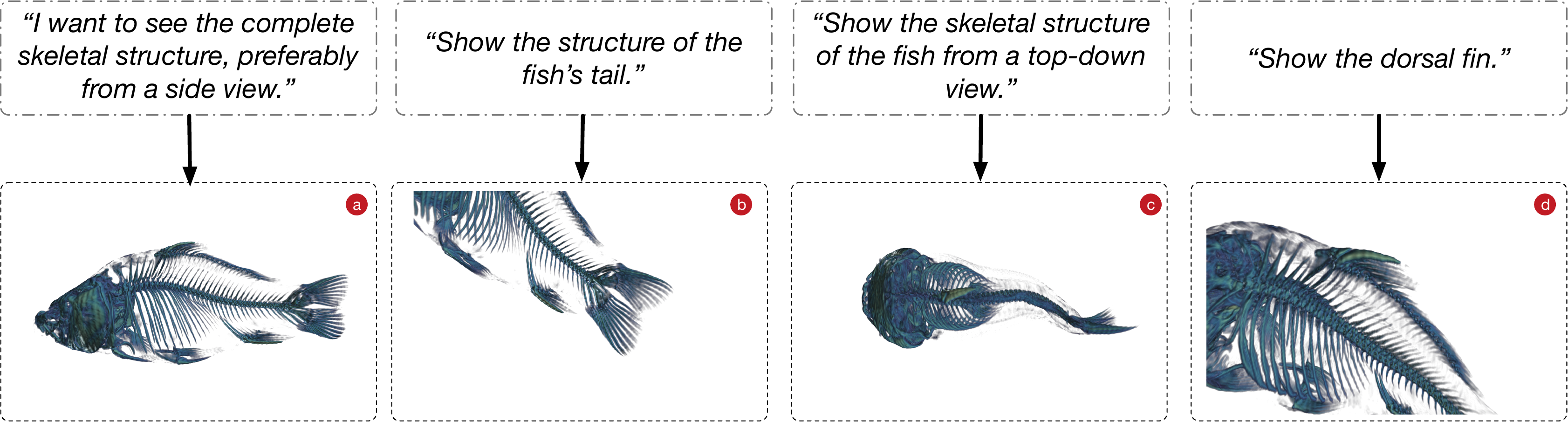}
    \caption{Case study demonstrating natural language-driven viewpoint navigation on the Carp Fish dataset. The user provides four language instructions to explore different anatomical structures of the fish: (a) full skeletal structure from a side view, (b) tail structure, (c) top-down view of the skeleton, and (d) dorsal fin. The system dynamically adjusts the camera viewpoint to satisfy each instruction, enabling intuitive, fine-grained exploration of volumetric data.}
    \label{Case1}
\end{figure*}

\begin{figure*}[th]
    \centering
    \includegraphics[width=1.0\textwidth]{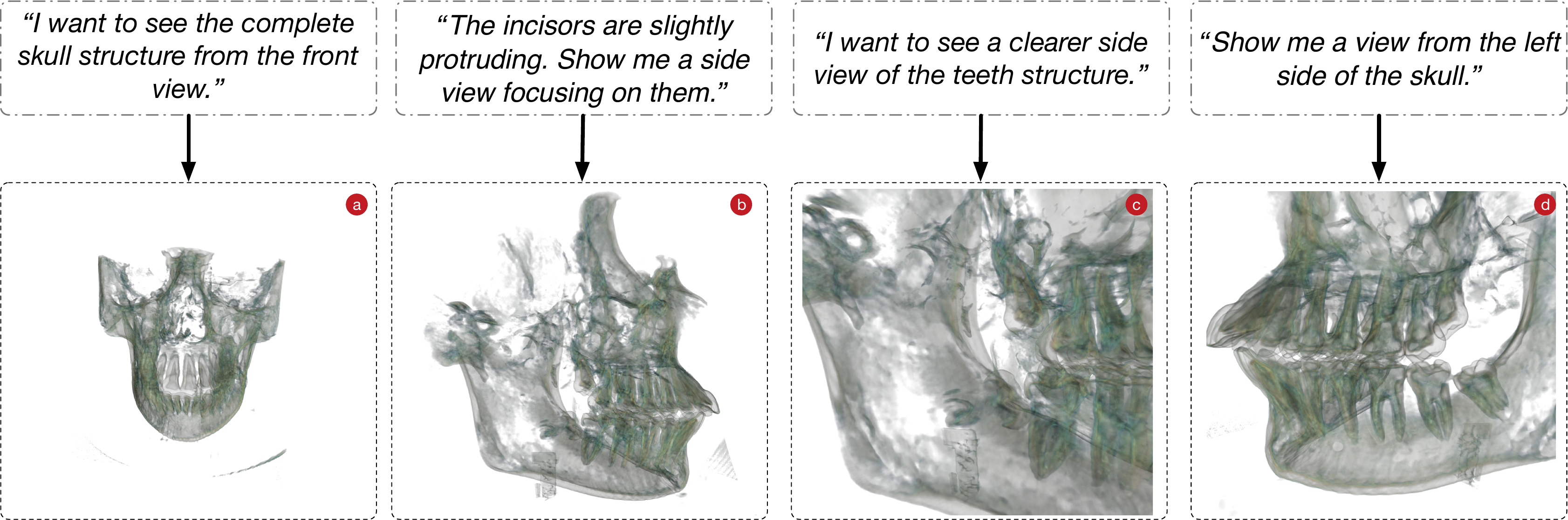} 
    \caption{Case study showcasing semantic viewpoint selection for human skull exploration. The system responds to user inputs focusing on dental and cranial features: (a) complete frontal skull view, (b) side view emphasizing protruding incisors, (c) clearer side view of the teeth structure, and (d) profile view from the left side. The viewpoint selection accurately captures user intent, facilitating targeted inspection of anatomical regions.}
    \label{case2}
\end{figure*}

\textbf{Impact of Fine-Tuning on CLIP Performance.}
To assess the effectiveness of fine-tuning CLIP on volumetric data, we evaluated its performance on three datasets before and after fine-tuning. As summarized in Table~\ref{tab:clip_finetuning_scores}, the fine-tuned model consistently outperforms the original CLIP, demonstrating significantly improved alignment between volumetric renderings and their corresponding textual descriptions.
\hot{
The evaluation is conducted on an unseen set of viewpoint–prompt pairs excluded from CLIP fine-tuning, ensuring that the results reflect generalization rather than memorization.
}
The Carp Fish dataset exhibited the most substantial gain, with the CLIP Score increasing from 18.16 to 28.32. The Skull dataset also showed notable enhancement, rising from 20.70 to 26.01. Even in the more challenging Argon Bubble dataset, the score improved from 13.49 to 23.42, highlighting the robustness of the fine-tuning process across diverse volumetric domains.
\hot{The textual prompts used for evaluation are sampled from a separate query set distinct from the fine-tuning corpus.}
These results validate the effectiveness of our fine-tuning strategy in adapting CLIP to the unique characteristics of volumetric data. By refining the model’s visual-textual alignment capabilities, our approach significantly enhances its ability to understand semantic content in volume-rendered images, ultimately improving viewpoint selection accuracy and overall user interaction quality.

\begin{figure*}[h]
    \centering
    \includegraphics[width=1.0\textwidth]{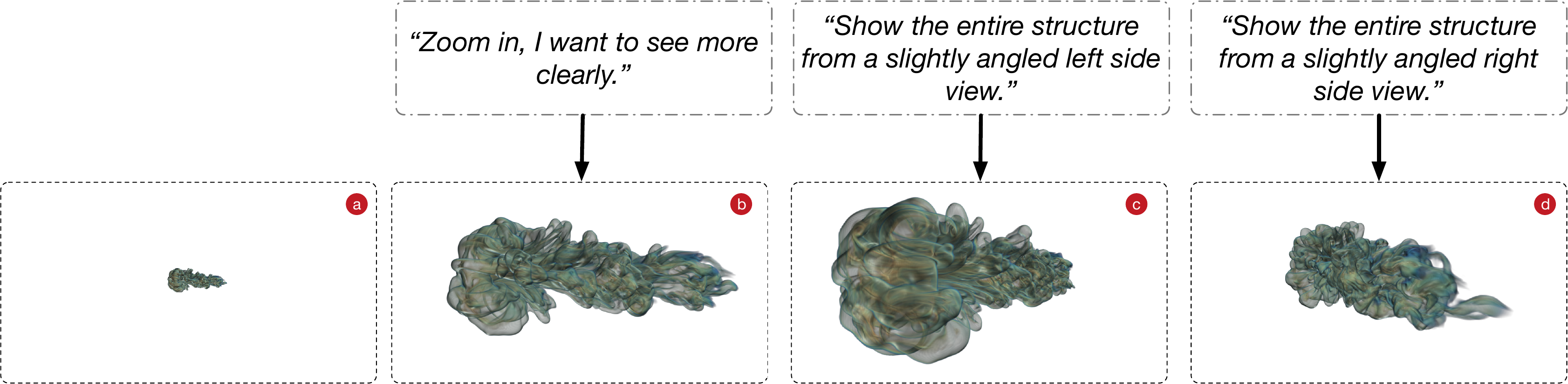} 
    \caption{Case study on volumetric navigation of the Argon Bubble dataset using natural language queries. Starting from a zoomed-out view, the user gradually refines the perspective to reveal structural details: (a) initial distant view, (b) zoom-in for clarity, (c) slightly angled left-side view, and (d) slightly angled right-side view. The system adjusts the camera accordingly to support symmetrical comparison and localized analysis.}
    \label{case3}
\end{figure*}

\subsection{Case Evaluation}
To further evaluate the practical capabilities of our framework, we present three case studies that highlight its effectiveness in navigating and interpreting volumetric datasets through natural language instructions. These scenarios illustrate how the system translates user-provided textual queries into semantically aligned viewpoints, thereby facilitating intuitive and targeted exploration of complex 3D structures. The selected cases span diverse domains: biological anatomy, medical imaging, and fluid simulation, demonstrating the generalizability and adaptability of our approach in different scientific visualization tasks.

\textbf{Case 1: Exploring the Skeletal Structure of a Fish.}
This case study uses a CT scan dataset of a fish to demonstrate how the proposed framework enables intuitive navigation through complex anatomical structures using natural language queries. The system interprets user instructions and dynamically selects semantically aligned viewpoints to facilitate fine-grained exploration.

The user begins by requesting a comprehensive view of the fish’s skeletal structure: 

\begin{quote} ``I want to see the complete skeletal structure, preferably from a side view." \end{quote}



In response, the system positions the camera to reveal the entire bone structure from a lateral perspective, clearly presenting the spine and fin rays. As shown in Fig.~\ref{Case1}(a), this viewpoint provides an unobstructed overview of the fish’s skeletal layout.

The user then refines their focus to a specific region with the query:

\begin{quote} ``Show the structure of the fish’s tail." \end{quote}



The framework adjusts the viewpoint to highlight the caudal fin area. The resulting view, shown in Fig.~\ref{Case1}(b), exposes the vertebral column’s extension into the tail and the intricate arrangement of supporting fin rays.

Next, the user asks to observe the structure from above:

\begin{quote} ``Show the skeletal structure of the fish from a top-down view." \end{quote}



The system reorients the camera directly overhead, capturing the spinal alignment and fin structures from a dorsal perspective (Fig.~\ref{Case1}(c)). This helps reveal the curvature and spatial organization of the bones across the body.

Finally, the user directs attention to the dorsal fin:

\begin{quote} ``Show the dorsal fin." \end{quote}
The camera is repositioned to emphasize the dorsal region. As depicted in Fig.~\ref{Case1}(d), the dorsal fin’s bony supports are clearly visible in relation to the spinal column.
This case illustrates how the system enables precise, intent-driven viewpoint selection through natural language input. Without requiring manual navigation or expert domain knowledge, users can explore volumetric data in a structured, semantically meaningful manner.



\textbf{Case 2: Detailed Exploration of a Human Skull.}


In this case study, we apply our framework to a volumetric CT scan of a human skull to demonstrate its effectiveness in supporting anatomical analysis through natural language interaction. The system interprets user instructions and dynamically adjusts viewpoints to reveal relevant cranial structures, enabling precise and targeted inspection.

The exploration begins with a query requesting an overall view of the skull from the front:

\begin{quote} ``I want to see the complete skull structure from the front view." \end{quote}


In response, the system selects a frontal viewpoint that clearly displays the facial bones, eye sockets, and nasal cavity. As shown in Fig.~\ref{case2}(a), this view provides a comprehensive perspective of the skull’s morphology, serving as a baseline for further exploration.

The user then refines the focus to a dental feature, specifically the protruding incisors:

\begin{quote} ``The incisors are slightly protruding. Show me a side view focusing on them." \end{quote}



To accommodate this request, the system adjusts the viewpoint to the lateral side of the skull. As illustrated in Fig.~\ref{case2}(b), the repositioned view highlights the upper and lower incisors, improving visibility of the dental alignment and aiding in the assessment of anatomical variations or potential abnormalities.

Next, the user requests a clearer side view of the overall dentition:

\begin{quote} ``I want to see a clearer side view of the teeth structure." \end{quote}



The camera is reoriented to maximize the visibility of the full dental arch, including incisors, canines, and molars. The resulting view (Fig.~\ref{case2}(c)) offers an unobstructed perspective on tooth arrangement and occlusion, which is particularly valuable for applications in orthodontics and forensic analysis.

Finally, the user requests a complete lateral view of the skull:

\begin{quote} ``Show me a view from the left side of the skull." \end{quote}

The system responds by rotating the viewpoint to the skull’s left profile. As shown in Fig.~\ref{case2}(d), the updated view clearly reveals important anatomical features such as the temporal bone, zygomatic arch, and mandible, offering insights into craniofacial structure and symmetry.

This case demonstrates the capability of the proposed method to facilitate detailed anatomical inspection through semantically guided viewpoint selection. By interpreting natural language descriptions, the system allows users to navigate complex volumetric data without requiring manual adjustment or expert knowledge in 3D navigation. The approach improves usability and interpretability, making it particularly beneficial in clinical, educational, and forensic settings.



\textbf{Case 3: Exploration of an Argon Bubble Dataset.}



In this case study, we demonstrate the adaptability of our framework by applying it to a volumetric dataset of an argon bubble. This example highlights how users can interactively refine viewpoints through natural language input to support detailed structural analysis.

The exploration begins with a request to enhance the visibility of the object:

\begin{quote} ``Zoom in, I want to see more clearly." \end{quote}



In response, the system zooms in on the bubble, generating a magnified view that reveals finer structural features. As shown in Fig.~\ref{case3}(a), this close-up allows users to better perceive the internal complexity and subtle surface details of the volume.

Next, the user seeks a comprehensive view of the bubble from a slightly angled left side perspective:

\begin{quote}  ``Show the entire structure from a slightly angled left-side view." \end{quote}



The system repositions the viewpoint accordingly, presenting the bubble from the specified angle. This view (\cref{case3}(b)) emphasizes the overall contour and flow of the structure, enabling a more holistic understanding of its geometry.

To compare the opposite side, the user issues a follow-up request:

\begin{quote} ``Show the entire structure from a slightly angled right side view." \end{quote}



The camera is shifted to the corresponding viewpoint, as illustrated in Fig.~\ref{case3}(c). This complementary perspective facilitates symmetry analysis and supports the identification of any irregularities in the structure, contributing to a more balanced assessment.

This case illustrates the system’s ability to support flexible, user-guided exploration of complex volumetric data. By allowing fine-grained control over camera orientation through natural language interaction, the framework enhances both accessibility and analytical depth, making it well-suited for applications in scientific visualization, simulation analysis, and educational contexts.

\begin{figure}[h]
  \centering
  \begin{subfigure}[t]{0.3\linewidth}
    \includegraphics[width=\linewidth]{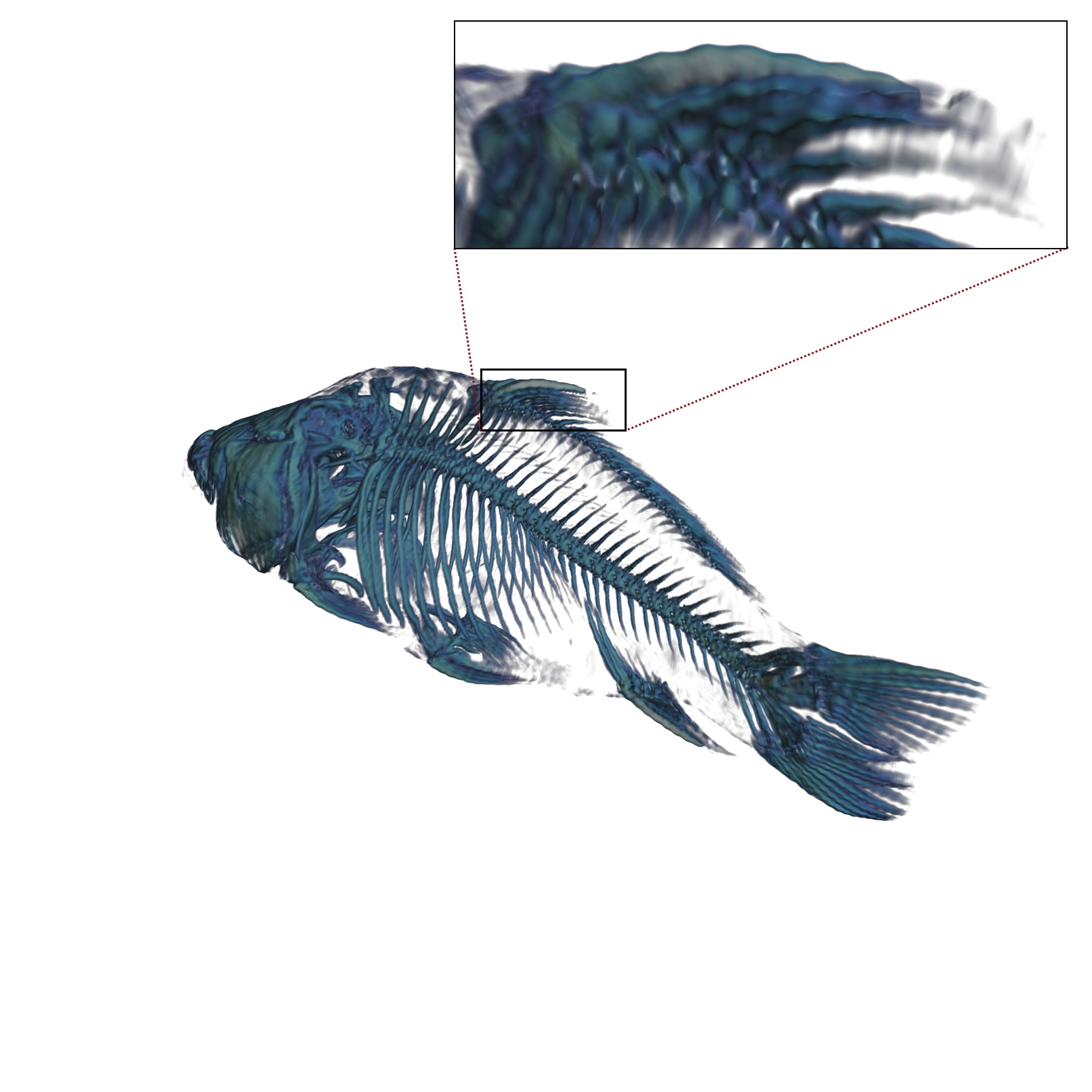}
    \caption{$128 \times 128 \times 256$}
  \end{subfigure}
  \hfill
  \begin{subfigure}[t]{0.3\linewidth}
    \includegraphics[width=\linewidth]{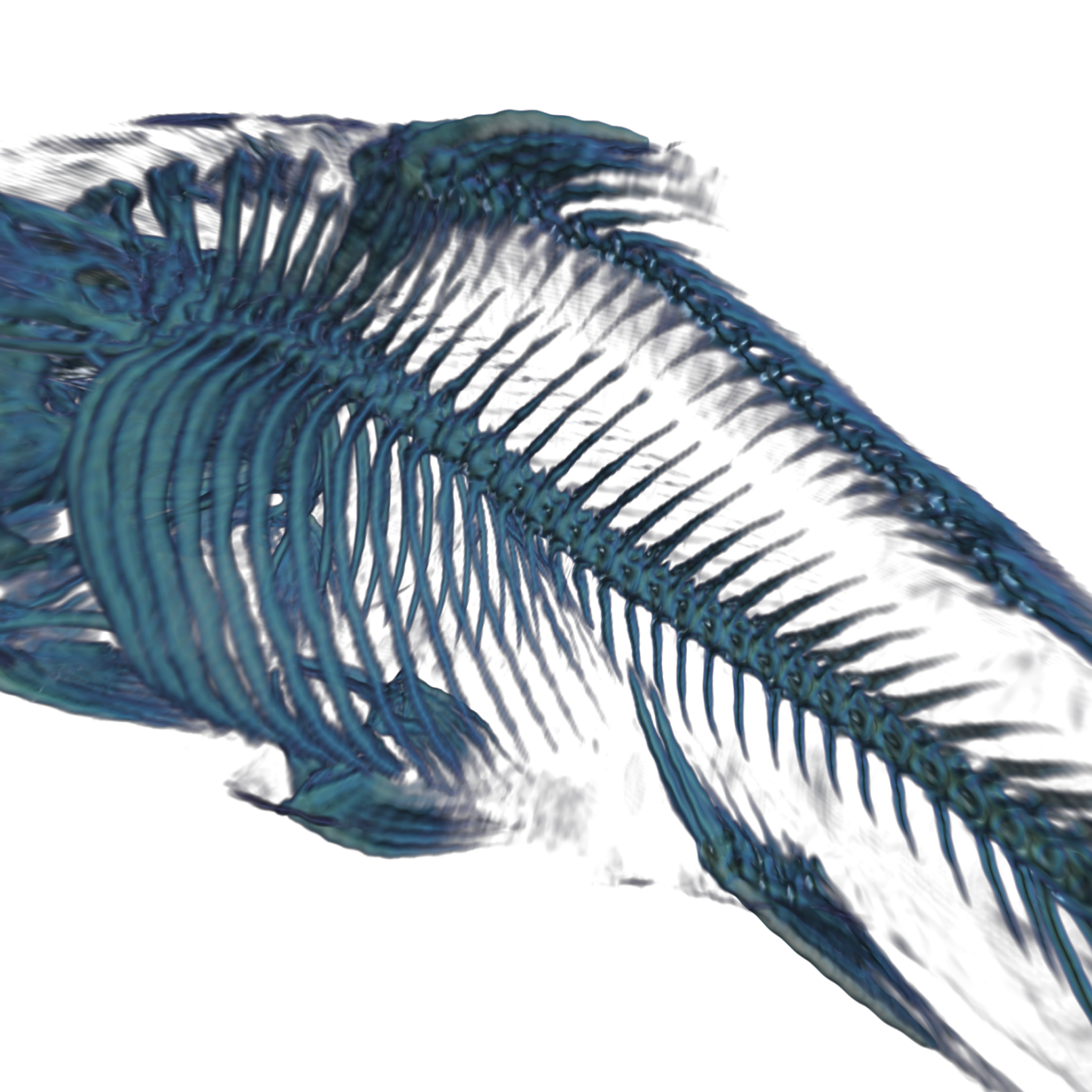}
    \caption{$64 \times 64 \times 128$}
  \end{subfigure}
  \hfill
  \begin{subfigure}[t]{0.3\linewidth}
    \includegraphics[width=\linewidth]{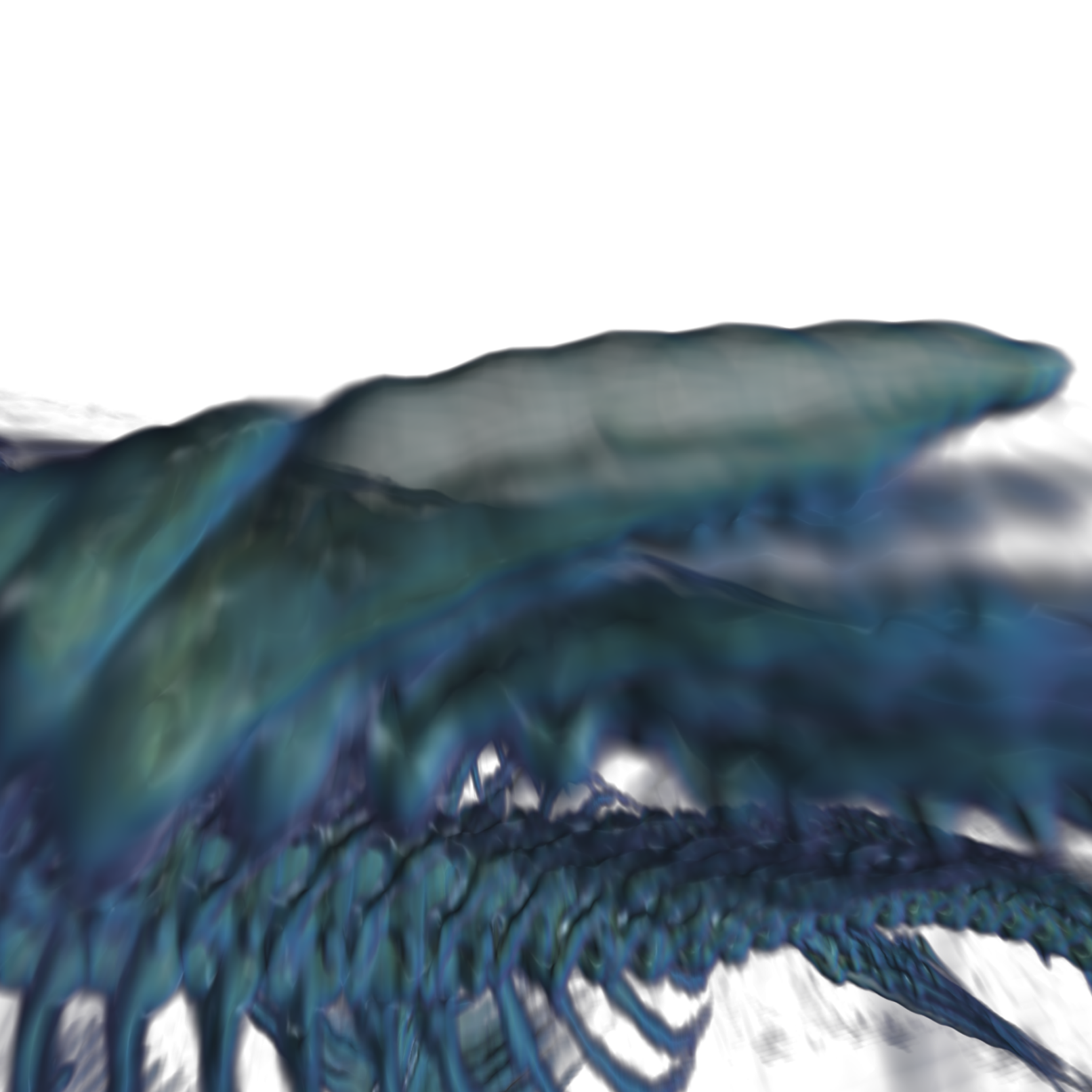}
    \caption{$32 \times 32 \times 64$}
  \end{subfigure}

  \vspace{1mm}

  \begin{subfigure}[t]{0.3\linewidth}
    \includegraphics[width=\linewidth]{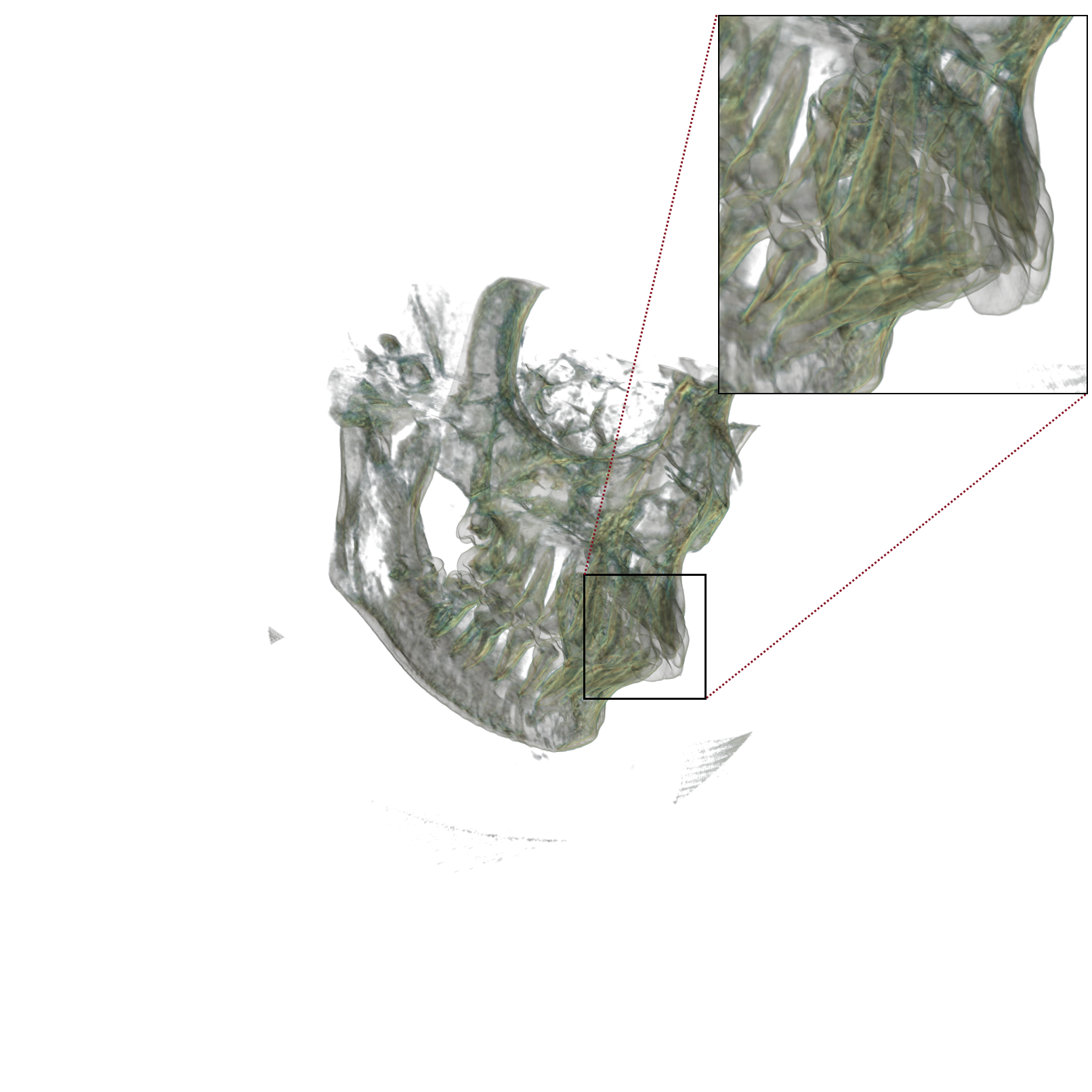}
    \caption{$128 \times 128 \times 128$}
  \end{subfigure}
  \hfill
  \begin{subfigure}[t]{0.3\linewidth}
    \includegraphics[width=\linewidth]{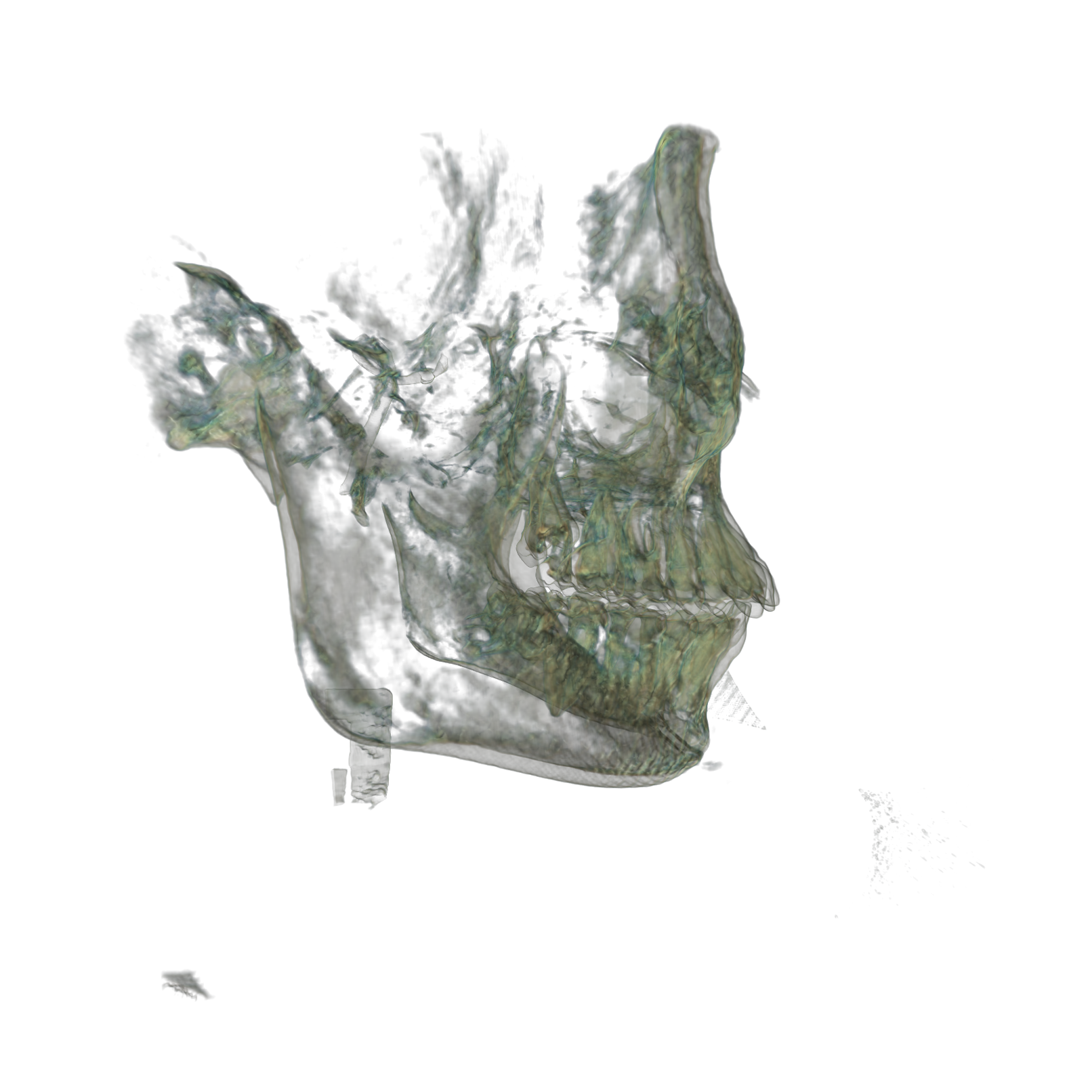}
    \caption{$64 \times 64 \times 64$}
  \end{subfigure}
  \hfill
  \begin{subfigure}[t]{0.3\linewidth}
    \includegraphics[width=\linewidth]{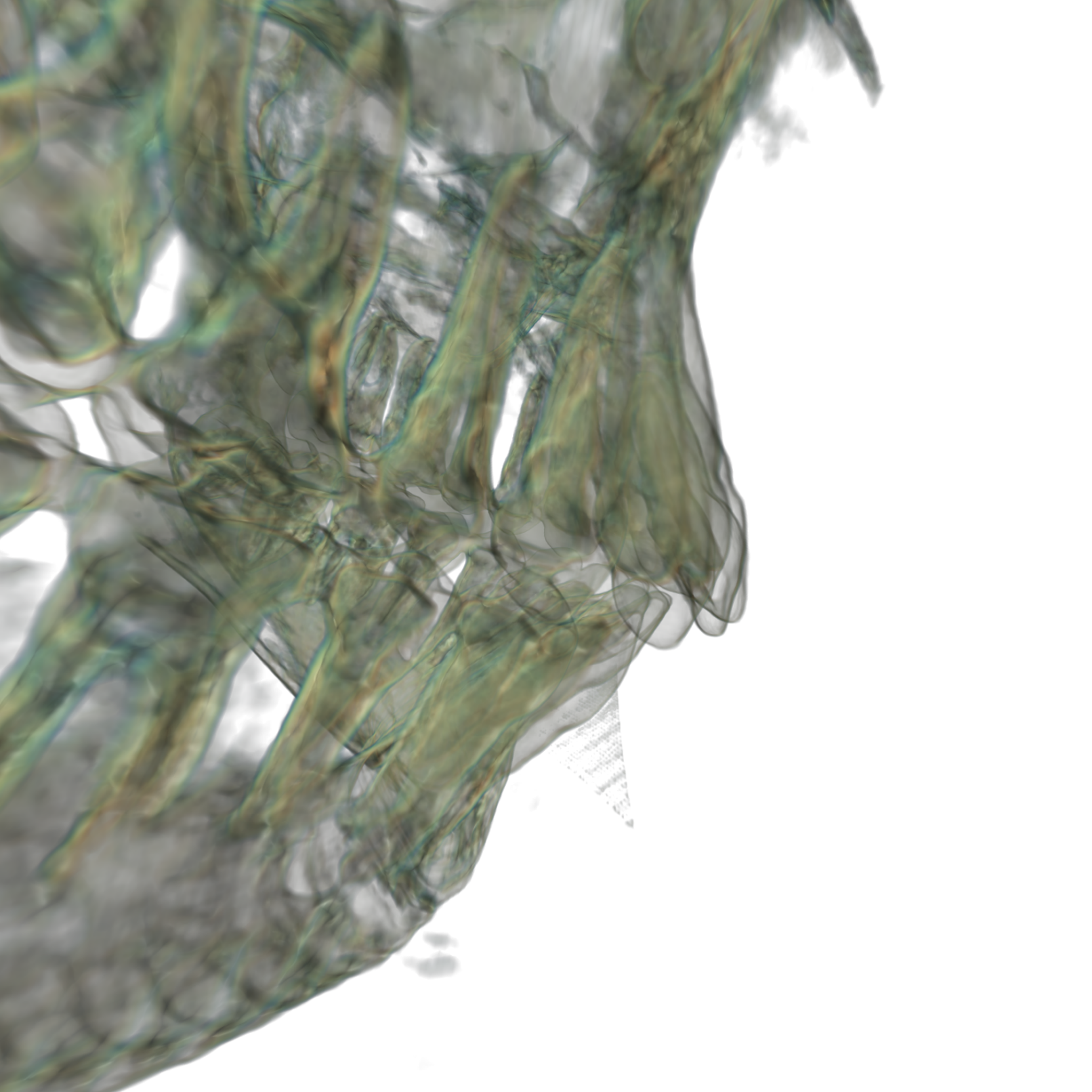}
    \caption{$32 \times 32 \times 64$}
  \end{subfigure}

  \vspace{1mm}

  \begin{subfigure}[t]{0.3\linewidth}
    \includegraphics[width=\linewidth]{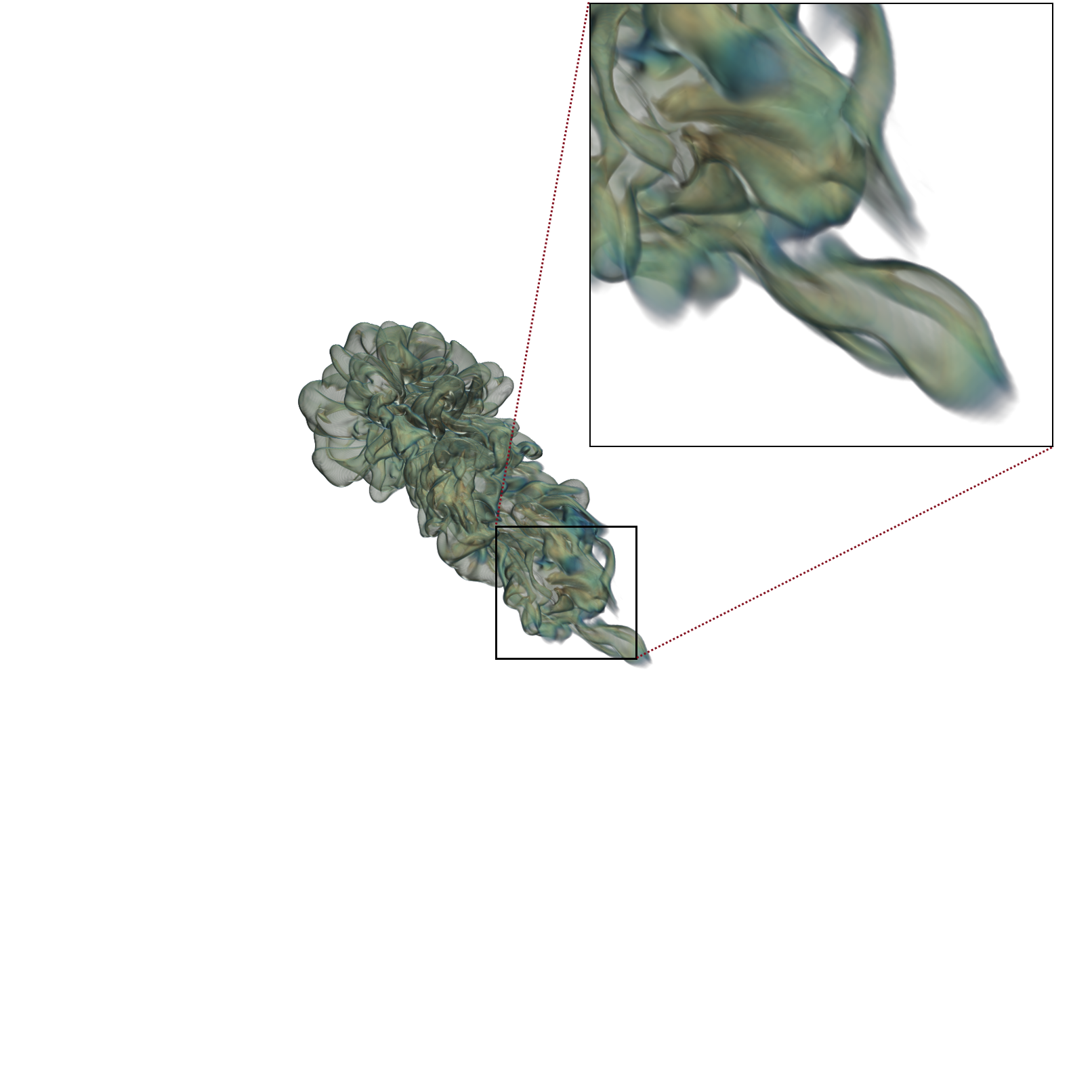}
    \caption{$128 \times 128 \times 320$}
  \end{subfigure}
  \hfill
  \begin{subfigure}[t]{0.3\linewidth}
    \includegraphics[width=\linewidth]{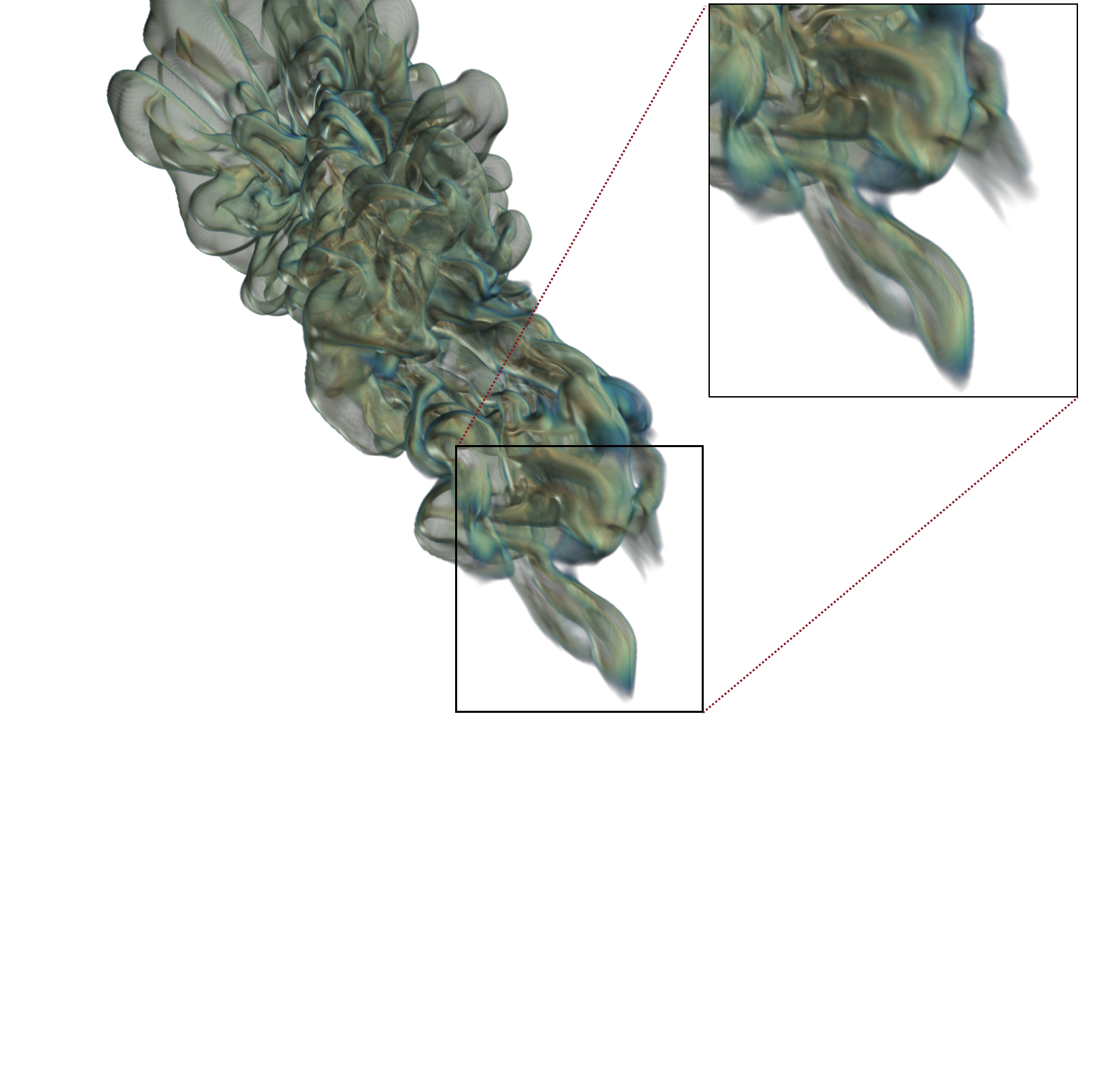}
    \caption{$64 \times 64 \times 160$}
  \end{subfigure}
  \hfill
  \begin{subfigure}[t]{0.3\linewidth}
    \includegraphics[width=\linewidth]{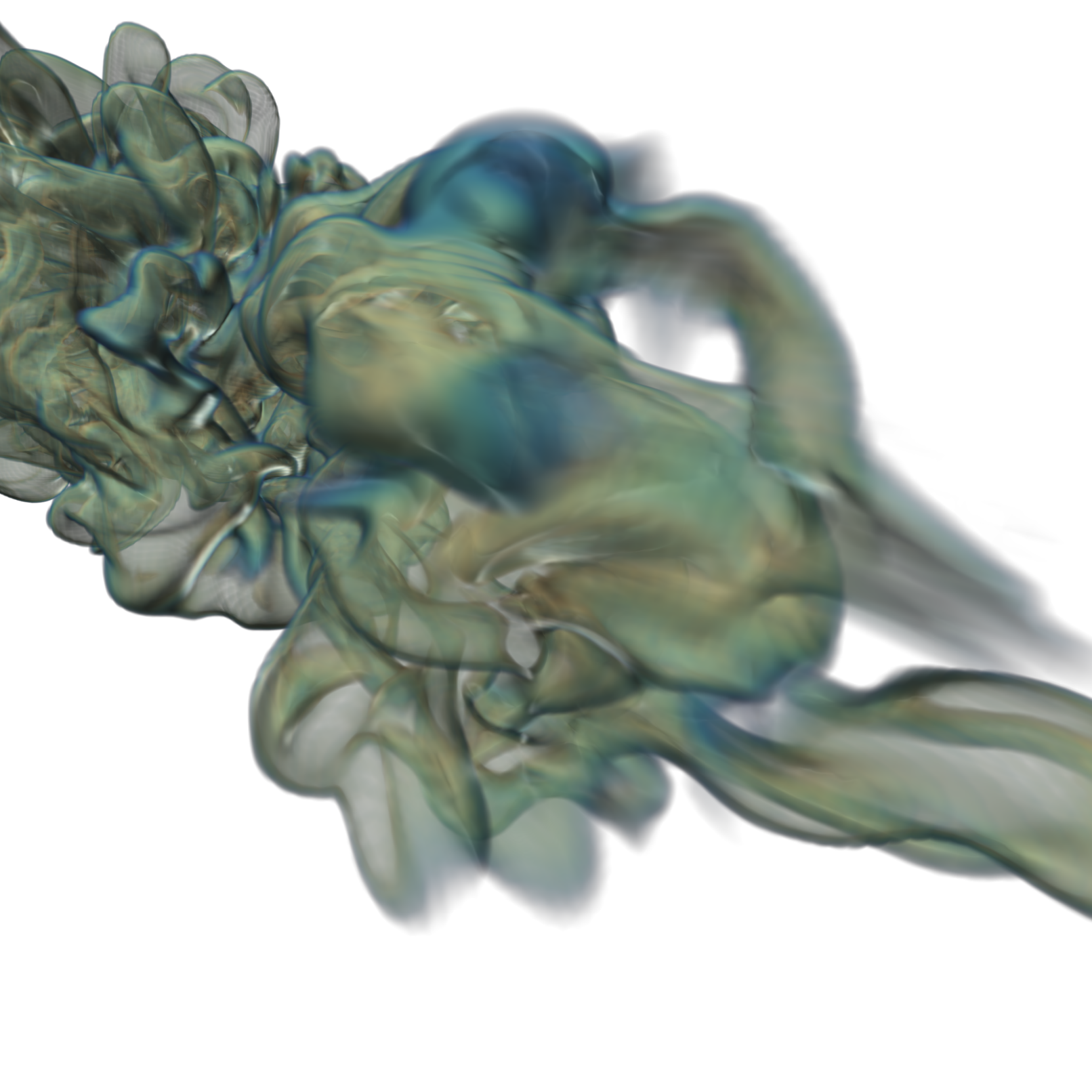}
    \caption{$32 \times 32 \times 80$}
  \end{subfigure}

  \caption{\hot{Qualitative ablation results comparing different volumetric block sizes. Columns represent block configurations: coarse (8 blocks), intermediate (64 blocks), and fine (512 blocks) from left to right. The block resolution is indicated in the corresponding subfigure caption. From top to bottom, the rows correspond to the Carp Fish, Skull, and Argon Bubble datasets, with prompts: ``Show the dorsal fin of the fish'', ``Side view of protruding front teeth'', and ``Show the turbulent wake region of the bubble.'', respectively.} }
  \label{fig:block-ablation}
\end{figure}

\hot{
\subsection{Block Size Sensitivity Analysis}
To assess the impact of block granularity, we conduct an ablation study by varying the number of blocks used during viewpoint optimization. Specifically, we compare three configurations: coarse partitioning with $2\times 2\times 2$ blocks, medium partitioning with $4\times 4\times 4$ blocks (our default setting), and fine partitioning with $8\times 8\times 8$ blocks. For each configuration, we evaluate the results on the same natural language queries across the Carp Fish, Skull, and Argon Bubble datasets.

As illustrated in \cref{fig:block-ablation}, all three block configurations produce optimized viewpoints that consistently focus on the queried structures within the volume, indicating a shared semantic understanding of the target features. However, the granularity of the partitioning significantly affects the scale of attention.

Coarse partitioning (a, d, g) results in large spatial coverage per block, leading to coarse-grained semantic representations. Optimized viewpoints in this setting tend to show entire global structures instead of focusing on local structures. For example, in the Skull dataset, while the optimized view (d) does fulfill the prompt ``side view of protruding front teeth", it presents a broader view of the skull. As a result, an additional zoom-in operation may be needed to reach the desired level of detail. In contrast, fine partitioning allows the model to focus on highly localized features such as the dorsal fin’s curvature, protruding teeth, or subtle bubble interfaces, as seen in (c, f, i). While this enables more detailed semantic alignment, it loses some context and introduces higher computational cost due to the increased amount of blocks. The intermediate configuration achieves a favorable trade-off, capturing task-relevant substructures with sufficient detail while preserving spatial coherence. 

These findings confirm that semantic meaning can be captured across a range of block resolutions, demonstrating the robustness of the representation. Nevertheless, block size remains an important hyperparameter, as it influences the efficiency and scale of viewpoint selection.
}

\section{Discussion}

\hot{
This work demonstrates that images can serve as an effective medium to connect natural language with 3D volumetric structures. This is valuable for cross-modal processing in scientific visualization, because 3D volumes lacks the large-scale labeled resources available in natural image domains. By leveraging pretrained vision-language models such as CLIP on rendered views, we are able to transfer semantic priors from image–text data into the volumetric domain. Our framework also shows the idea of interpreting volumes through block-wise representations, is effective for machine-level understanding as well.

Looking ahead, we aim to support more interactive and temporally coherent tasks such as conversational exploration and continuous camera navigation. However, the following challenges need to be tackled:

\textbf{Varying feature scales.} Features of different spatial scales often appear along a camera path. Some structures require global context, while others demand fine local detail. Our current use of fixed-resolution blocks can lead to either loss of detail or unnecessary fragmentation. Addressing this will likely require more adaptive representations, possibly by combining multiscale blocks with geometric structures such as isosurfaces to better capture semantic variation.

\textbf{Incorporating human feedback.} Planning a sequence of viewpoints based on multi-turn conversation or long-form prompts is more complicated than static view selection. While reinforcement learning has proven useful in similar ordering problems~\cite{yu2025von}, planning exploratory paths in volumetric spaces involves reasoning over spatial relationships and semantic priorities among features. In contrast to view selection, we lack large-scale trajectory-labeled training data or even intermediate medium, making fully supervised learning difficult. Human feedback may offer crucial guidance, but also raises the challenge of learning effectively from vague, sparse, or subjective signals.

\textbf{Recognizing dynamic focuses.} The inverse task of semantic narration from viewpoint sequences poses the challenge of identifying dynamic features in animations. Scientific features are often scattered or subtle, making it difficult to determine what should be described at each frame based on local appearance alone. Effective narration requires understanding the entire trajectory to infer the shifting focus and organize observations into coherent, structured language. This demands global spatiotemporal reasoning across visualization, 3D structures, and semantic domains.
}

\section{Conclusion}

\comp{We present a framework for natural language-driven viewpoint selection in volumetric data exploration, which incorporates a semantic block representation scheme that utilizes images as a medium to learn semantic information from volumetric blocks. By capturing the semantics of these blocks, the framework provides valuable guidance for optimizing viewpoints. This approach integrates pretrained vision-language models, such as CLIP, with reinforcement learning to dynamically adjust viewpoints based on semantic rewards. Experimental results demonstrate the effectiveness of our method in capturing user intent, improving navigation efficiency, and enhancing interpretability, with case studies illustrating its ability to refine viewpoints in complex anatomical and geometric scenarios.}

\hot{
Building on this foundation, future work will explore more expressive architectures for volumetric block encoding. In particular, attention-based encoders and adaptive sampling strategies may improve the representational fidelity and semantic granularity of block features. These enhancements are expected to further strengthen the alignment between natural language instructions and 3D structural content, contributing to more robust and effective volumetric data exploration. In addition, addressing the broader challenges discussed (e.g., multi-scale semantic reasoning, human-in-the-loop learning, and temporal focus recognition) will be essential for advancing the cross-modal understanding of complex volumetric data.}

\section*{Acknowledgements}
This research was supported in part by the National Natural Science Foundation of China through grants 62172456 and 62372484.
The authors would like to thank the anonymous reviewers for their insightful comments.

\bibliographystyle{abbrv-doi-hyperref}
\bibliography{template}

\clearpage

\appendix

\pagestyle{fancy}
\fancyhead{} 
\fancyhead[L]{Natural Language-Driven Viewpoint Navigation for Volume Exploration via Semantic Block Representation}
\fancyhead[R]{Xuan Zhao, and Jun Tao} 

\setcounter{page}{1}
\setcounter{figure}{0}
\setcounter{section}{0}
\setcounter{table}{0}

\renewcommand{\thefigure}{S\arabic{figure}}
\renewcommand{\thetable}{S\arabic{table}}

\begin{figure}[ht]
\centering
\begin{subfigure}[t]{0.48\linewidth}
    \centering
    \includegraphics[width=\linewidth]{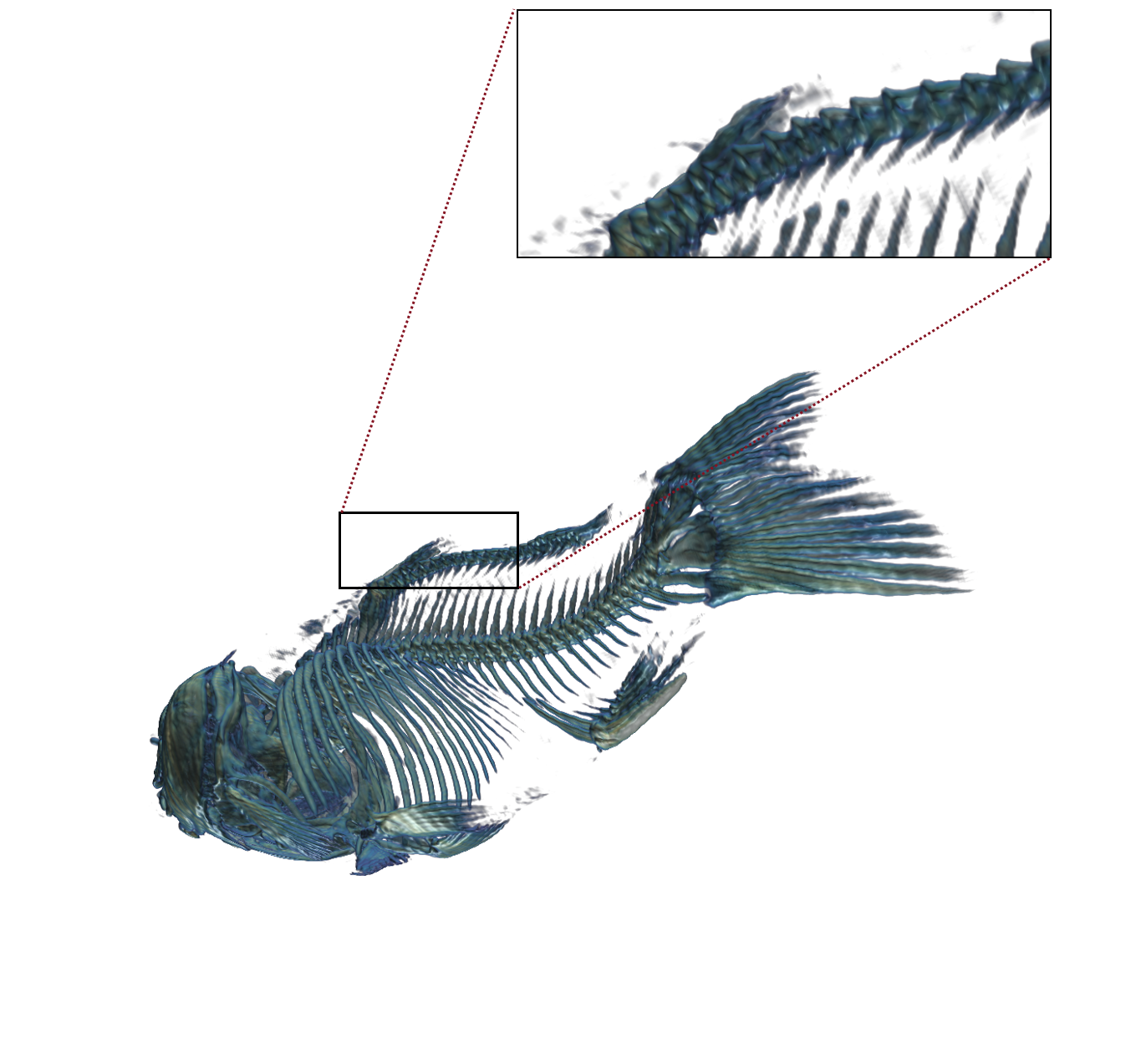}
    \caption{Prompt: \textit{``Show me the dorsal\\fin of the fish.''}}
    \label{fig:failure_bubble_vorticity}
\end{subfigure}
\begin{subfigure}[t]{0.48\linewidth}
    \centering
    \includegraphics[width=\linewidth]{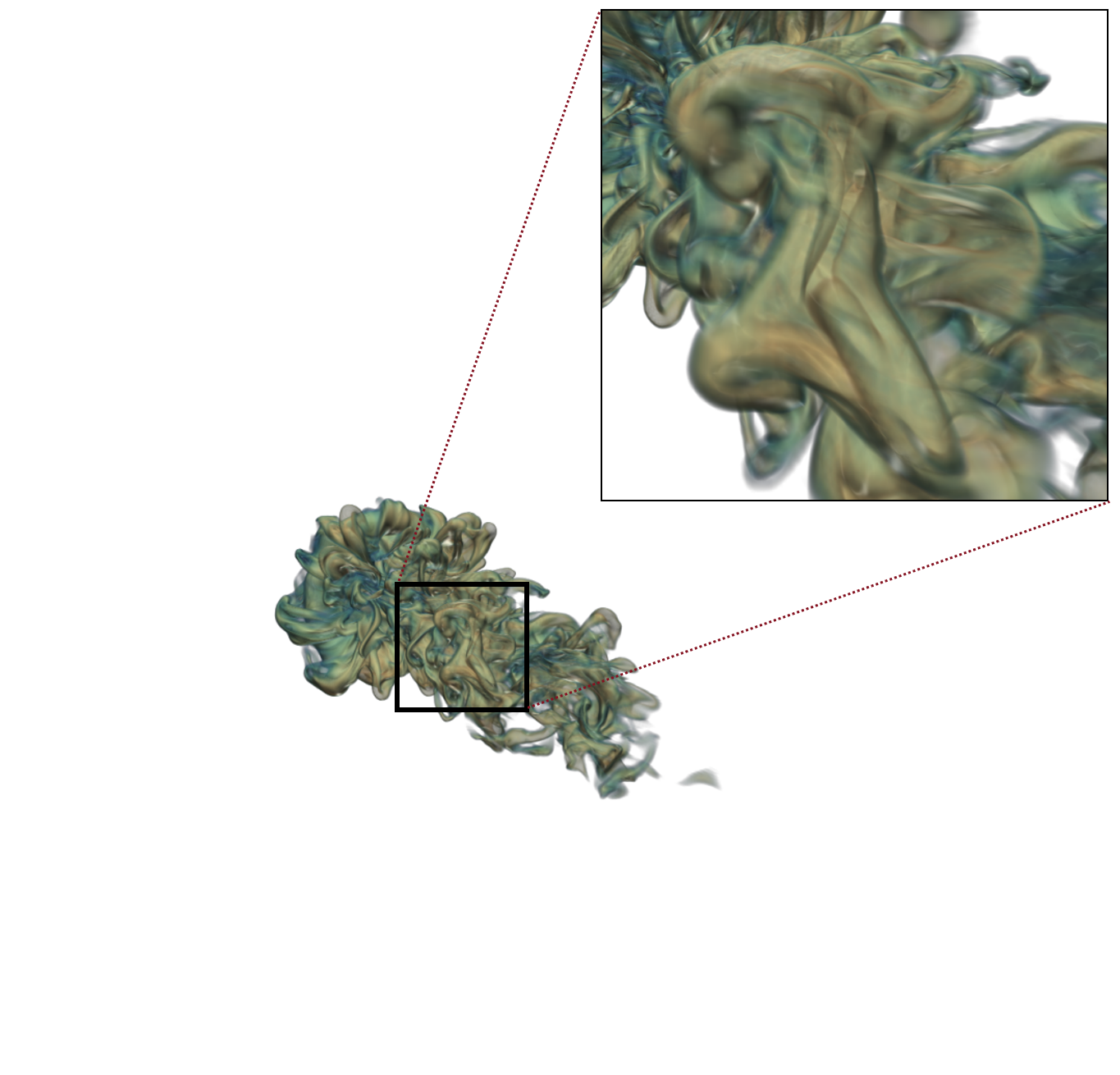}
    \caption{Prompt: \textit{``Zoom in to observe\\the bubble's outer surface in detail.''}}
    \label{fig:failure_bubble_origin}
\end{subfigure}
\hfill
\caption{\hot{Failure cases on the Argon Bubble and Carp fish dataset.}}
\label{fig:failure_cases}
\end{figure}

\hot{
\section{Failure cases}
Despite the overall effectiveness of our method, we observe several failure cases where the system responds to the prompts but fails to fully satisfy the intended semantics, either due to partial occlusion of the target or difficulty in resolving conflicting instructions. Fig.~\ref{fig:failure_cases} presents two representative examples. 

\textbf{Partial occlusion.} In (a), the prompt \textit{``Show me the dorsal fin of the fish''} leads to a viewpoint that partially reveals the dorsal fin. However, the fin is not fully exposed and is partially occluded by the body, which reduces its saliency. This illustrates a challenge in resolving self-occlusion without sufficient semantic cues.

\textbf{Conflicting instructions.} In (b), the prompt \textit{``Zoom in to observe the bubble’s outer surface in detail''} presents an inherent conflict: it requests both a zoomed-in view and coverage of a large, surface-level structure that spans the entire volume. The system finally selects a view showing the entire volume, but fails to deliver sufficient magnification or surface detail, illustrating a limitation in handling conflicting spatial semantics within a single prompt.
}

\begin{figure*}[htbp]
    \centering
    \begin{minipage}{0.33\linewidth}
        \centering
        \includegraphics[width=\linewidth]{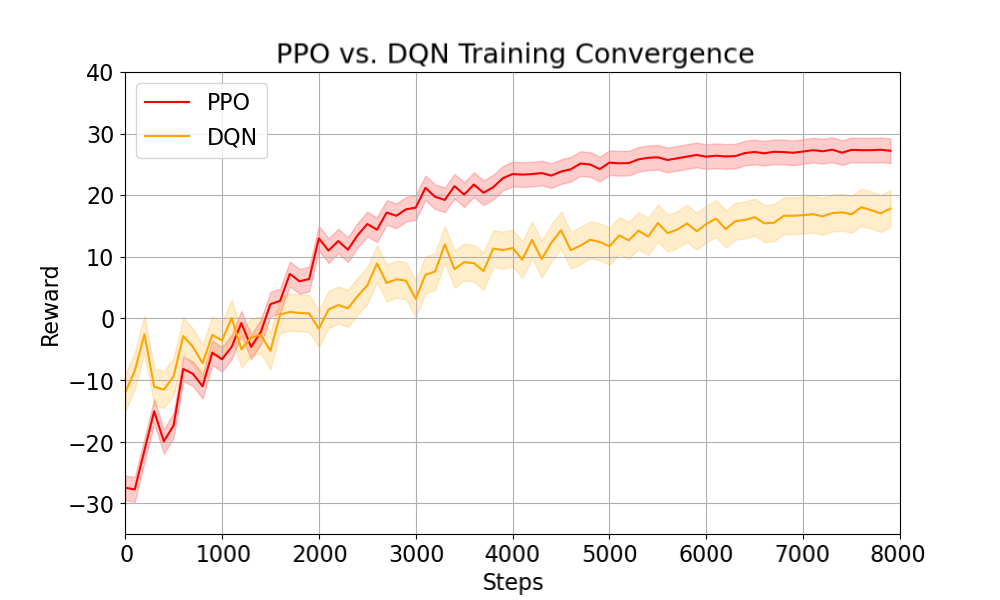}\par
        (a) Carp Fish
    \end{minipage}
    \hfill
    \begin{minipage}{0.33\linewidth}
        \centering
        \includegraphics[width=\linewidth]{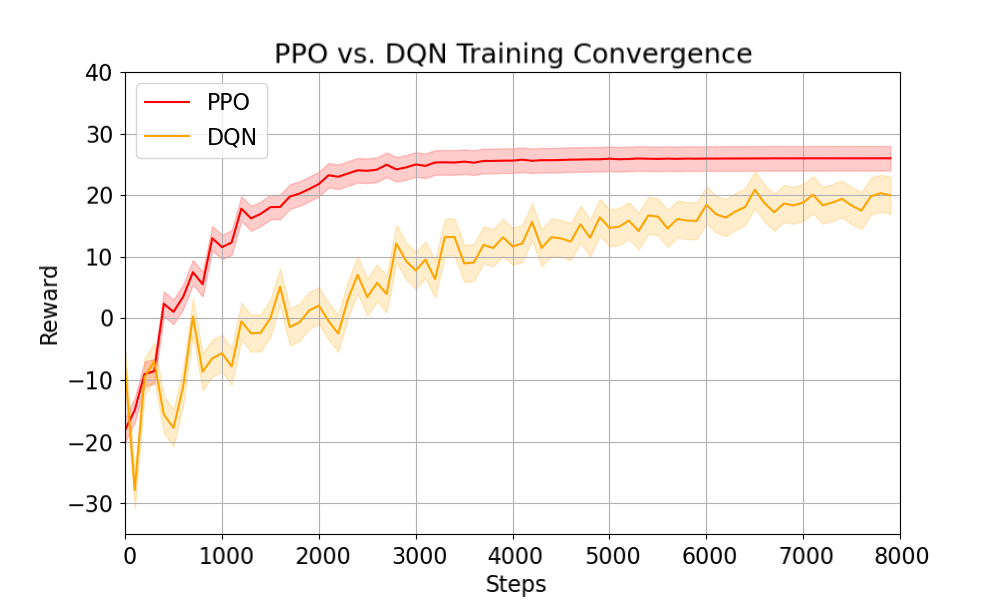}\par
        (b) Skull
    \end{minipage}
    \hfill
    \begin{minipage}{0.33\linewidth}
        \centering
        \includegraphics[width=\linewidth]{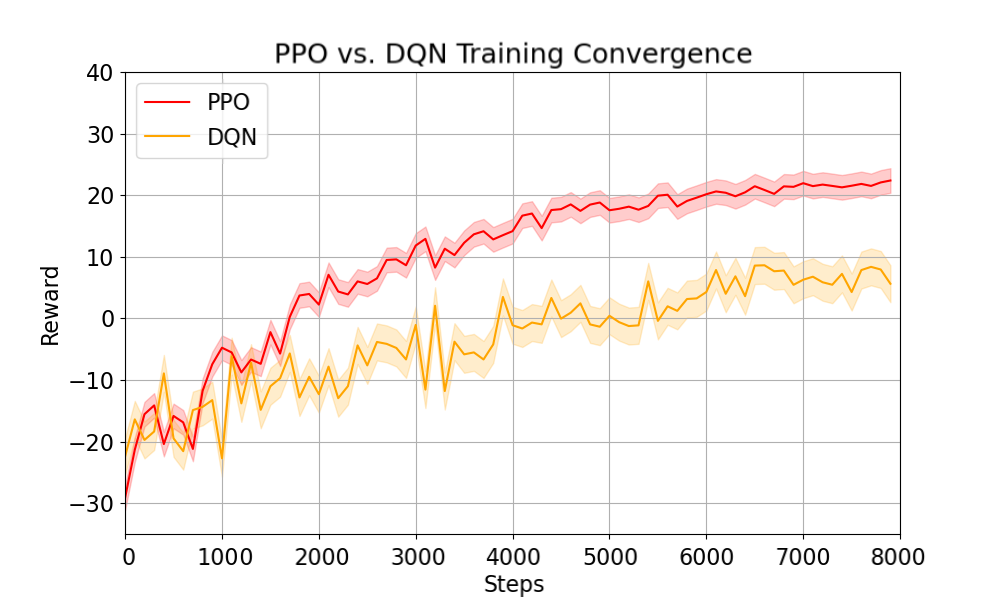}\par
        (c) Argon Bubble
    \end{minipage}
    \caption{Comparison of PPO and DQN in training convergence across datasets. Each plot reports the evolution of average reward over training steps. For all three datasets, PPO yields higher semantic alignment rewards with faster stabilization.}
    \label{ppo and dqn}
\end{figure*}

\hot{
\section{Impact of optimization method}
We choose proximal policy optimization (PPO) for identifying optimal viewpoints. PPO employs clipped policy gradients that stabilizes training and enables smoother updates. However, we should note that our approach does not rely on specific optimization methods. To validate our choice, we compare PPO against a popular alternative, Deep Q-Network (DQN).

Fig.~\ref{ppo and dqn} shows the training curves of PPO and DQN across the three datasets. For all three datasets, PPO consistently outperforms DQN in both convergence speed and final CLIP Score. For instance, in the Carp Fish dataset, while PPO starts from a lower score, it finally reaches a score of 28 compared to DQN's 17; in the Skull dataset, PPO scores 26 versus DQN's 20. Even in the more challenging Argon Bubble dataset, where both algorithms exhibit early fluctuations, PPO ultimately achieves better semantic alignment through more effective viewpoint selection.

These results highlight the advantages of PPO's policy-gradient approach, which enables stable updates and efficient exploration in high-dimensional spaces. In contrast, DQN's reliance on value estimation limits its adaptability, leading to slower convergence and suboptimal performance. The consistent superiority of PPO provides strong empirical support for our framework’s design in reinforcement learning-driven volumetric exploration.
}

\hot{
\section{Viewpoint Prompt Construction Guidelines}
To support intuitive interaction, our system allows users to specify viewpoint search objectives using natural language prompts. While the language interface is flexible and robust to variation, we find that certain prompting patterns are more effective than others in producing semantically aligned and spatially accurate viewpoints. This appendix provides guidance for constructing such prompts based on our empirical findings and the underlying model design.

An effective viewpoint prompt typically consists of two elements: (1) a reference to a specific anatomical or structural target within the volume (e.g., ``dorsal fin'', ``rib cage'', ``upper incisors''), and (2) an indication of the desired view direction or framing (e.g., ``from the front'', ``side view'', ``zoom in''). Prompts that include both components provide the clearest supervision for semantic alignment and spatial disambiguation. For example, ``Show the dorsal fin of the fish from a side view'' is generally more successful than ``Show the structure", which lacks a clear spatial or semantic anchor.

We encourage users to use concise and descriptive phrases that clearly state what to look at and how to look at it. 
The use of anatomical or visually identifiable terms, such as ``tail structure'', ``bubble surface'', or ``dental arch'' helps the model ground the instruction in interpretable visual features. Spatial modifiers such as ``frontal'', ``top down'', or ``zoom in'' help the agent distinguish between different perspectives of the same structure, which is especially important when occlusion or symmetry is present.
Moreover, prompts that are overly abstract (e.g., ``complex topology'') or vague (e.g., ``the interesting part'') may lead to inconsistent or suboptimal results. Since the system currently handles single-turn instructions, multi step queries (e.g., ``First show the skull, then focus on the nose'') are not supported.
In summary, users can achieve the best results by writing short, direct prompts that name the target structure and specify the intended view.

\section{Prompts for Generating Text Descriptions}
To facilitate the generation of descriptive image-text pairs across different types of volumetric datasets, we constructed tailored sets of prompt instructions for each dataset. These prompts were designed to guide a large language model (LLM) in annotating rendered images with viewpoint-sensitive and semantically meaningful descriptions. Specifically, the prompts for the Carp Fish dataset focus on anatomical structures and spatial orientation, the Skull dataset emphasizes skeletal landmarks and viewing angles, and the Argon Bubble dataset targets flow patterns, boundary complexity, and structural features derived from fluid dynamics. Tables~\ref{tab:fish_instruction_prompts},~\ref{tab:skull_instruction_prompts}, and~\ref{tab:argon_instruction_prompts} present the complete sets of prompts used for each dataset. These prompts were crafted to balance global summaries and local feature analysis, ensuring semantic diversity and expressiveness in the generated annotations.

\section{Impact of Viewpoint Sampling Strategies}
To evaluate the impact of different viewpoint sampling strategies on semantic alignment, we compare three configurations: (1) \textbf{Volume-Centered Only}, where all candidate viewpoints are sampled from a fixed-radius sphere and oriented toward the volume center; (2) \textbf{Block-Centered Only}, where viewpoints are generated to focus on randomly selected blocks; and (3) \textbf{Hybrid (Ours)}, which combines both types to enrich candidate diversity. All other components, including the encoder, reward function, and training procedure, are kept identical.

We conduct this experiment on the Skull dataset using a set of five natural language prompts previously used in our qualitative analyses:
\textit{``Show a frontal view focusing on the upper incisors.''},
\textit{``I want to see the complete skull structure from the front view.''},
\textit{``The incisors are slightly protruding. Show me a side view focusing on them.''},
\textit{``I want to see a clearer side view of the teeth structure.''},
\textit{``Show me a view from the left side of the skull.''}.
Table~\ref{tab:sampling_strategy_ablation} reports the average CLIP Score across these prompts for each sampling strategy.

\begin{table}[htbp]
\centering
\caption{\hot{Impact of different viewpoint candidate strategies on CLIP alignment.}}
\begin{tabular}{l c c}
\toprule
\textbf{Sampling Strategy} & \textbf{CLIP Score ↑}  \\
\midrule
Volume-Centered Only & 23.87  \\
Block-Centered Only  & 25.11  \\
Hybrid (Ours)        & \textbf{29.76}  \\
\bottomrule
\end{tabular}
\label{tab:sampling_strategy_ablation}
\end{table}

As shown in Table~\ref{tab:sampling_strategy_ablation}, relying solely on volume-centered sampling yields the lowest CLIP Score, likely due to insufficient coverage of off-center anatomical structures. Block-centered sampling improves semantic alignment by enabling finer spatial control, but tends to produce redundant or unstable views without global context. Our hybrid strategy achieves the best performance, by providing a more balanced candidate pool that captures both global and local structural information. 
These results validate our design choice of combining volume-centered and block-centered sampling to enhance viewpoint optimization.
}







\begin{figure}[h]
\centering
\includegraphics[width=\linewidth]{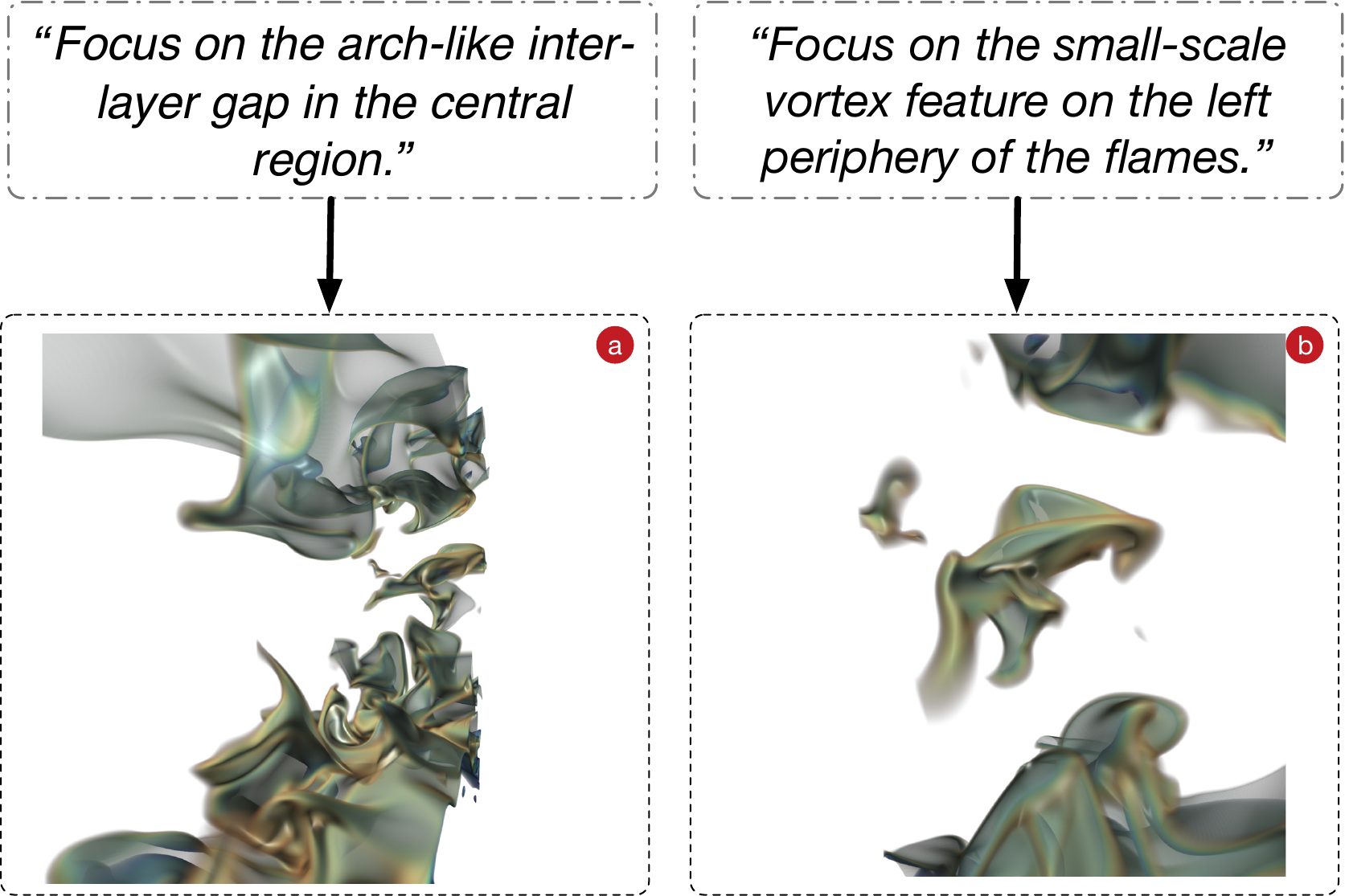}
\caption{\hot{
Exploring the combustion dataset with semantic viewpoint selection.(a) focuses on the arch-like inter-layer gap in the central region, revealing spatial separation and geometric complexity of the folded scalar layers. (b) focus on the small-scale vortex feature on the left periphery of the flames, highlighting localized vortical structures critical for turbulent mixing analysis.
}
}
\label{fig:combustion_case}
\end{figure}

\hot{
\section{Additional Qualitative Case Study}
In this case study, we demonstrate the applicability of our framework to the \textit{Combustion} dataset, derived from direct numerical simulation of temporally evolving turbulent non-premixed flames. The simulation generates five scalar variables: heat release (HR), mixture fraction (MF), vorticity (VORT), mass fraction of hydroxyl radical (YOH), and scalar dissipation rate (CHI). For this study, we use the mixture fraction attribute, which represents the scalar field of fuel-oxidizer mixing. The dataset dimensions are \(480\times 720\times 120\).

We start by focusing on a distinct spatial gap within the volume:

\begin{quote}
``Focus on the arch-like inter-layer gap in the central region.''
\end{quote}

As shown in Fig.~\ref{fig:combustion_case}(a), the system centers the viewpoint on the arch-shaped gap formed between folded scalar layers in the central region. This perspective emphasizes the geometric complexity and spatial separation of layered flame structures, facilitating detailed semantic exploration.

Next, the prompt directs attention to a fine-scale flow feature:

\begin{quote}
``Focus on the small-scale vortex feature on the left periphery of the flames.''
\end{quote}

In response, the system shifts the camera to highlight the localized vortical structure near the left boundary of the volume, as illustrated in \cref{fig:combustion_case}~(b). This view reveals intricate flow perturbations and localized scalar variations essential for analyzing turbulent mixing dynamics.

This case study demonstrates the ability of our natural language-driven viewpoint selection framework to navigate features in simulation data, providing targeted views that capture both larger-scale structural gaps and fine-scale vortical phenomena in turbulent combustion data.}

\begin{table*}[htbp]
\centering
\caption{\hot{Textual Description Instructions for LLM on the Carp Fish Dataset}}
\begin{tabular}{p{16cm}}
\toprule
\textbf{Prompt Instructions} \\
\midrule
Describe the fish’s anatomical structures and their spatial arrangement as observed in this CT volume. \\
Analyze the spatial orientation of the fish in the rendered view. Use anatomical directions (e.g., left/right, top/bottom) to locate the head and tail. \\
Explain the internal skeletal and external anatomical features of the fish that are visible in this volumetric rendering. \\
Provide a structured and comprehensive description of the fish’s morphology as depicted in this 3D scientific visualization. \\
Identify the head, body, and tail regions of the fish and describe their orientation relative to the viewing perspective. \\
Describe the fish's appearance, including its skeletal structure and notable features, based on the current viewpoint. \\
Provide a general overview of the fish’s overall structure as seen from a distance. \\
Describe what level of detail is visible in this rendering of the fish. \\
What part of the fish appears to be the focus of this image? \\
Describe the observable structures and their relative prominence from this view. \\
Does this image appear to be taken from a close or distant viewpoint? Justify your answer with visible features. \\
Estimate how zoomed-in this view is. Then describe the key features accordingly. \\
If any part of the fish is occluded or unclear in this view, describe what is missing or hard to observe. \\
Which anatomical parts are most clearly visible in this image, and which are partially hidden? \\
\bottomrule
\end{tabular}
\label{tab:fish_instruction_prompts}
\end{table*}

\begin{table*}[htbp]
\centering
\caption{\hot{Textual Description Instructions for LLM on the Skull Dataset}}
\begin{tabular}{p{16cm}}
\toprule
\textbf{Prompt Instructions} \\
\midrule
Describe the anatomical features of the human skull visible in this volumetric rendering, including bones and cavities. \\
Analyze the orientation of the skull in this view. Use anatomical terms (e.g., anterior/posterior, left/right) to describe the facing direction. \\
Identify prominent skeletal features such as the mandible, maxilla, zygomatic arch, and orbital cavities from this perspective. \\
Provide a structured explanation of the skull's morphology as seen in this X-ray volume scan. \\
Discuss the clarity and visibility of key bone structures (e.g., jaw, teeth, nasal cavity) in this rendering. \\
Is the skull viewed from a frontal, lateral, or oblique angle? \\
Estimate the proximity of the camera to the skull. Does the viewpoint suggest a close-up inspection or an overview? \\
What specific part of the skull appears to be emphasized or in focus in this view? \\
Describe the spatial relationship between the upper and lower jaw bones in this rendering. \\
Are any anatomical regions partially occluded or indistinct? Describe what is missing or less visible. \\
Which features are rendered most prominently in this volumetric image, and which are less pronounced? \\
\bottomrule
\end{tabular}
\label{tab:skull_instruction_prompts}
\end{table*}

\begin{table*}[htbp]
\centering
\caption{\hot{Textual Description Instructions for LLM on the Argon Bubble Dataset}}
\begin{tabular}{p{16cm}}
\toprule
\textbf{Prompt Instructions} \\
\midrule
Describe the overall flow structure and identify any dominant directional features present in this rendering. \\
Analyze the boundary or contour-like regions in this image. Which parts appear more defined or abrupt? \\
Identify the front (head-like) and rear (tail-like) regions of the bubble and describe their differences in shape and detail. \\
Explain the complexity and layering observed in the surface of the bubble from this viewpoint. \\
Discuss whether the viewpoint emphasizes large-scale shape or fine-grained turbulence. \\
Describe the outer shell or boundary layer of the volume and how it contrasts with the inner regions. \\
From this rendering, estimate if the viewpoint is aligned with or against the main flow direction. \\
Describe any observable vortex structures or instabilities visible from this angle. \\
Which part of the flow structure seems closest to the viewer? Which regions appear to recede into the volume? \\
Estimate the spatial scale shown in the view. Are we observing a zoomed-in detail or an overview of the entire volume? \\
Which flow regions are clearly defined, and which are partially transparent or occluded? \\
What are the main visual features in focus in this rendering? Are there strong edges, gradients, or smooth transitions? \\
\bottomrule
\end{tabular}
\label{tab:argon_instruction_prompts}
\end{table*}

\end{document}